\documentclass{article}

\usepackage[final]{corl_2022} 

\usepackage{algorithm}
\usepackage[noend]{algorithmic}
\usepackage{amsmath}
\usepackage{amssymb}
\usepackage{color}
\usepackage{xcolor}
\usepackage{graphicx}
\usepackage{dblfloatfix}
\usepackage{multicol}
\usepackage{multirow}
\usepackage[numbers]{natbib}
\usepackage{subcaption}
\usepackage{tabularx}
\usepackage{times}
\usepackage{setspace}
\usepackage{makecell}
\usepackage{siunitx}
\usepackage{enumitem}
\usepackage{xurl}
\usepackage{mathtools}

\usepackage{hyperref}
\hypersetup{
    colorlinks=true,
    linkcolor=orange,
    filecolor=magenta,      
    urlcolor=orange,
    citecolor=orange,
}

\usepackage{titlesec}

\newcommand{\REV}[1]{#1}

\title{PLATO: Predicting Latent Affordances \\ Through Object-Centric Play}
\author{Suneel Belkhale, Dorsa Sadigh}

\begin{document}
\titlespacing\section{0pt}{8pt plus 2pt minus 2pt}{0pt plus 2pt minus 2pt}

\thispagestyle{plain}
\pagestyle{plain}

\maketitle

\begin{abstract}
Constructing a diverse repertoire of manipulation skills in a scalable fashion remains an unsolved challenge in robotics. One way to address this challenge is with unstructured human play, where humans operate freely in an environment to reach unspecified goals. Play is a simple and cheap method for collecting diverse user demonstrations with broad state and goal coverage over an environment. Due to this diverse coverage, existing approaches for learning from play are more robust to online policy deviations from the offline data distribution. However, these methods often struggle to learn under scene variation and on challenging manipulation primitives, due in part to improperly associating complex behaviors to the scene changes they induce. Our insight is that an object-centric view of play data can help link human behaviors and the resulting changes in the environment, and thus improve multi-task policy learning. In this work, we construct a latent space to model object \textit{affordances} -- properties of an object that define its uses -- in the environment, and then learn a policy to achieve the desired affordances. By modeling and predicting the desired affordance across variable horizon tasks, our method, Predicting Latent Affordances Through Object-Centric Play (PLATO), outperforms existing methods on complex manipulation tasks in both 2D and 3D object manipulation simulation and real world environments for diverse types of interactions. Videos can be found on our \href{https://sites.google.com/view/plato-corl22/home}{website}.

\end{abstract}



\section{Introduction}
The field of robotics has seen tremendous progress in solving manipulation tasks, but learning a general multi-task policy
remains an open challenge. Imitation learning methods are sample-efficient at replicating demonstrated behaviors, but are often presented with structured, predefined tasks and therefore struggle to generalize outside of the data distribution~\cite{pomerleau1998alvinn, ross2010efficientreductions}.
Rather than using predefined task demonstrations, recent work has shown that learning from \textit{play} data -- an unstructured form of demonstration without predefined goals -- can lead to policies that are more robust to online deviations~\citep{lynch2020learning}. Play data is easy to collect at scale since it requires no task specification or manual resetting, and play can have broad data coverage over the set of object interactions necessary for performing a variety of tasks. 

Existing approaches for learning from play sample short horizon-length windows from play data to learn goal-conditioned imitation policies in an offline fashion
~\cite{lynch2020learning, gupta2019relay}. However, not all facets of the demonstrator's behavior are captured by the goal alone; thus prior work learns a latent space to model the variation in human behaviors. Such a latent space captures a representation of ``plans," for example plans representing reaching or grasping motions, for a given goal state during play. These latent plans can then help inform robot policies at test time~\cite{lynch2020learning}.

These approaches make several restrictive assumptions. Firstly, they assume that given a fixed short-horizon window of play, the agent's goal is the environment state at the end of the window. This assumption however is not always true: since the desired robot state is not available at test time, the goal is chosen as the environment state a few seconds in the future. Since the goal might be close or the same as the initial state, it can be uninformative for the policy, nor does it necessarily represent the human's true goal. 
Crediting behavior to an incorrect environment goal can obscure \textit{why} a human chose the actions during that window, and thus hinder policy learning. For example, if we want to grasp and move a block on a table
but our goal is sampled after the robot has initiated motion but before initiating the grasp, the sampled goal will have no change in the object state and thus gives us no information on the true goal that the user had in mind.
Secondly, by randomly sampling windows from play data, the learned latent space will be forced to model all sequences of behaviors \textit{equally}, even though many sequences will be less critical to achieving the desired environment goal.
These restrictive assumptions leads to suboptimal behaviors when increasing the complexity and variability of tasks, e.g., tasks with varying horizons, when learning from play. 


Instead of defining goals and plans based on arbitrary horizon lengths, 
we posit that humans often define goals and plan in terms of interactions with objects: rather than planning over the individual joint motions required to grasp and open a door handle, we think about turning the handle and then opening the door.
Our key insight is that viewing offline play data as diverse object interactions enables better modeling of both a human's goals and the behaviors that can achieve these goals.
Rather than learning to represent fixed short-horizon robot trajectories, or \textit{plans}, we bias the latent space towards learning demonstrated \textit{object affordances} that accomplish tasks in object-space. Imitation policies can then condition on these affordances directly to gain insight into human's behaviors.


We propose an algorithm, PLATO -- Predicting Latent Affordances Through Object-Centric Play -- that automatically segments play into a series of object interactions using proprioceptive information, and then learns a latent affordance space over object interactions. Simultaneously, it learns to imitate human actions over variable horizon interactions, conditioned on the latent affordance and goal.
%
By considering object interactions in play and correctly attributing goals to robot behaviors, PLATO builds a robust mapping between the true goals, desired object affordances, and actions on the robot over varying horizons.
This leads to PLATO significantly outperforming prior methods especially when increasing the complexity and variability of play data.
Our contributions are: 
\vspace{-0.1cm}
\begin{enumerate}[leftmargin=*]
    \item We have developed an object-centric paradigm for learning from unstructured human play data, which views play as sequential, unlabeled interactions with objects.
    \vspace{-0.1cm}
    \item We have developed a new algorithm, PLATO, that extracts and leverages these interactions to model diverse object affordances from play data and learn a robust policy. 
    \vspace{-0.1cm}
    \item We have tested PLATO on a number of 2D and 3D manipulation environments in simulation and the real world, including diverse objects such as blocks, mugs, cabinets, and drawers, with broad coverage over possible tasks and object affordances. We demonstrate that PLATO substantially outperforms state-of-the-art learning from play baselines in these complex manipulation tasks. 
\end{enumerate}

\section{Related Work}
In this section, we will discuss prior work in goal-conditioned imitation learning, learning from play, and object-centric policy learning. Our work brings an object-centric perspective on learning from play to learn effective imitation policies that generalize across different manipulation tasks.

\smallskip
\noindent\textbf{Imitation Learning.} 
Imitation Learning is a common method for learning robot policies from human demonstrations, where a policy learns to mimic human actions \cite{pomerleau1998alvinn}. These methods often struggle to generalize to new environments and to learn from multi-task data \cite{SCHAAL1999233, ARGALL2009469, ross2010efficientreductions}.
To improve \emph{generalization}, Ho, et al. introduced generative adversarial IL to learn from imitation data while matching the expert policy distribution \cite{ho2016generative}. Recently, implicit imitation learning policies using energy-based models have also been shown to improve generalization as compared to explicit policy models \cite{florence2021implicit}. 
To enable \emph{multi-task} learning, one can condition the policy on goal states, either explicitly labelled or inferred via hindsight experience replay \cite{codevilla2017drivingcil, ding2019goal, andrychowicz2017hindsight}. Other methods have leveraged meta-learning for multi-task imitation learning to enable better multi-task performance and one-shot generalization \cite{duan2017oneshot}. A key limitation of these works is the dependency on demonstration quality and quantity, for example the state-action coverage within the dataset for each task \cite{mandlekar2021matters}. Our work utilizes an approach built on top of goal-conditioned imitation learning to learn \emph{multi-task} policies. However, we use play data as opposed to expert demonstrations, which enables a much broader state-action coverage to learn \emph{generalizable} and robust policies.

\smallskip
\noindent \textbf{Learning from Play.} Human play data is defined as unstructured and unsupervised human interaction with an environment, as a means to let the user guide the data collection process, and provide broader coverage of the task-relevant state and goal space. Because humans can freely choose how to interact with the environment, play data is easy and cheap to collect without the need for prior task specification or environment resets.
These factors make play data a rich source for skill learning; as a result, imitation policies trained on play data are more robust to deviations from the expert trajectories than those trained on single-task demonstrations~\cite{lynch2020learning}. 
The state of the art method, Play-LMP, learns latent ``plans" over short, fixed horizon trajectories from play data~\cite{lynch2020learning}. This approach can also be used to ground language with play data using pretrained language models~\cite{lynch2020grounding}. These learned short-horizon skills have also been chained together to accomplish long horizon tasks using motion planning inspired techniques \cite{ichter2020broadlyexploring}. Additionally, using small variations in the horizon length of play windows has been shown to improve test time performance~\cite{gupta2019relay}. These methods benefit from the broad coverage of play data, but suffer from distribution shift at test time due to incorrectly inferred goals during training. Therefore, our approach also imitates play, but we take an object-centric view: by learning from diverse interactions with the environment, we can more accurately infer the human's goals during play and thus obtain a better state-action-goal distribution.

\smallskip
\noindent \textbf{Object-Centric Modeling and Policy Learning.}
There is strong evidence that humans have neural pathways specific to object recognition, dynamics understanding, and scene segmentation~\cite{GRILLSPECTOR2003159}. Inspired by humans, this notion of object centrism has been applied in many recent works in object manipulation for better policy generalization and sample efficiency. Formalisms like object-oriented MDPs have been introduced to better model \textit{effects} of agent behaviors as functions of individual objects and their affordances~\cite{diuk2008object}. Object affordances have also been learned from labeled interaction examples in multi-task environments and were shown to benefit policy learning downstream~\cite{nag2020affordance, khaz2021visaff}. Planning relative to object reference frames and primitives can enable plans to generalize to changing environments~\cite{toki2020, dalal2021accelerating}. Similarly, focusing on objects during policy learning can allow behaviors to generalize to novel scenarios \cite{devine2018objcentric, zhao2021consciousnessinspired}. A recent work leveraged labelled object motions in play data to predict grasp points, and showed RL becomes more sample efficient post-grasping~\cite{borja2022affordance}. We employ object centrism to guide learning from play data, specifically by extracting affordances through unsupervised temporal segmentation over object interactions. 

\section{Predicting Latent Affordances Through Object-Centric Play}
\label{sec:method}

A play dataset $D_{\text{play}}$ consists of $N$ varying length episodes of undirected, human generated state-action trajectories $T_j,\ \forall j \in \{1,\dots,N\}$, where $T_j = \{s_1, a_1, \dots , s_{L_j}\}$ for length $L_j$. The state feature space $S$ may be learned or predefined, and consists of a proprioceptive state space $S^r$ (robot state space) and an environmental state space $S^o$ (object state space) such that $S^r \oplus S^o = S$.  The goals that generated these trajectories are not included in the dataset, and are defined as $o_g\in G$, for goal space $G \subseteq S^o$. We use $o$ and $s^o$ interchangeably to refer to environmental state, and likewise for $r$ and $s^r$ to refer to the robot state. We assume that goals only consist of the environmental state, since access to proprioceptive state goals at test time is unrealistic as it requires the robot having access to a policy for achieving the given environmental goal.

Given access to this play dataset $D_{\text{play}}$, a new initial state $s \in S$, and an object goal state $o_g$, our problem is to learn a robot policy $\pi$ to achieve the desired goal state. 
In prior work, sampled trajectories of play $\tau$ consist of a contiguous fixed-horizon segment from an episode $T_j$. For simplicity we denote sampling these segments from play as $\tau \sim D_{\text{play}}$, where the length of the segment is the fixed horizon $H$.
We propose viewing play sequences from a bird's eye view, namely object interactions, instead of from myopic and fixed 1-2 second windows used in prior work. From this perspective, we hypothesize human play is just a series of \textit{environment interactions} induced by a robot's actions. If we can somehow detect where interactions begin and end, we can learn to relate interaction and pre-interaction robot behaviors to the state of the world post-interaction. First, we discuss how we detect interactions in the environment in offline play by leveraging proprioceptive cues. Next, we formalize how we properly choose and associate goals with robot behaviors.



\begin{figure*}[t!]
    \centering
    \includegraphics[width=\linewidth]{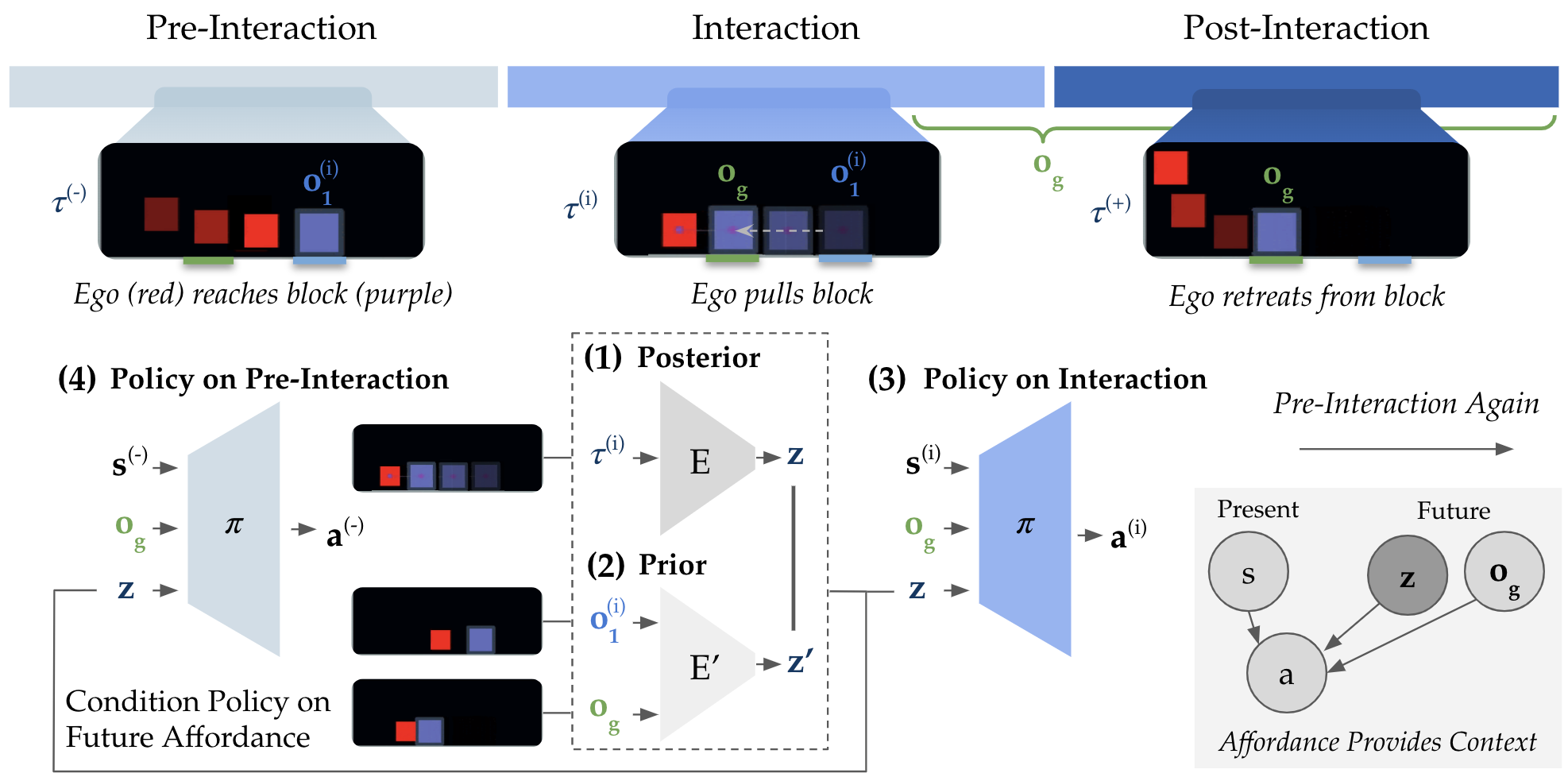}
    \caption{Our method shown on pre-interaction (purple), interaction (blue), and post-interaction (green) periods. (1) Posterior $E$ encodes the interaction sequence \smash{$\tau^{(i)}$} into affordance $z$. (2) Prior $E'$ encodes object start and goal states \smash{$o^{(i)}_1$} and $o_g$ to predict $z$. $o_g$ is sampled from post-interaction. (3) Policy trained to output actions on interaction period conditioned on affordance, and simultaneously (4) to output actions from pre-interaction period conditioned on the ``future" affordance. \REV{The assumed causal structure is shown in the lower right: PLATO claims the robot behavior can be explained with knowledge of the long term goal and future affordance $z$: the policy \textit{reasons} about its desired affordance and goal to determine its actions.}
    }
    \label{fig:clfp}
\end{figure*}



\smallskip\noindent\textbf{Segmenting Play into Interaction Phases}:
When we interact with an environment, we change its state through our own actions, whether it be through direct or indirect contact. In this work, we focus on single-object interactions, but we emphasize that interactions can be defined even over multiple objects that are being influenced by the robot behavior (e.g., in tool-mediated manipulation, interaction is defined between the tool and the environment). We break down an interaction into the following phases:
\vspace{-5pt}
\begin{enumerate}[leftmargin=*]
    \item \textbf{Pre-interaction:} This phase usually involves orienting the robot to interact with an object, e.g., reaching the purple block in Fig.~\ref{fig:clfp}.
    \item \textbf{Interaction:} This phase involves joint and interdependent motions between the robot and the object(s), e.g., pulling the purple block in Fig.~\ref{fig:clfp}.
    \item \textbf{Post-interaction:} This phase involves separation from the object and any downstream effects on the object, e.g., the purple block coming to rest after releasing it in Fig.~\ref{fig:clfp}. For repetitive interactions, this is the same as the next pre-interaction period.
\end{enumerate}
\vspace{-5pt}
To account for these periods, we \REV{detect the robot's \textit{influence} over the environment, e.g., grasping a door handle or lifting a mug, to automatically segment play into the periods above. For detecting single-object interactions,} contact signals are readily available with modern robots. The contact signal is smoothed and then chunks of play with contiguous contact are labelled as interaction periods. \REV{Several ideas for extending PLATO to handle multi-object interaction detection are discussed in detail in Appendix~\ref{app:limits}.}
Since play consists of many back-to-back interactions, the post interaction phase of one segment, e.g., retreating from a block and moving towards a cup, is simply the pre-interaction for the next segment, e.g., lifting the cup.
We denote pre, interaction, and post windows with superscripts $^{(-)}$, $^{(i)}$, and $^{(+)}$, respectively. 


\REV{\textbf{Sampling Task Relevant Goals Post-Interaction}: By segmenting play into interaction phases, we can more accurately sample an accurate goal environment state. Critically, goals \textit{result} from an interaction by definition. At the end of an interaction, the environment state will very likely have changed. Therefore, rather than arbitrarily labeling the goal as the last state of a 1-2 second window, as in prior work, we can label the goal as any state from the end of the interaction through post-interaction (see $o_g$ in Figure~\ref{fig:clfp}). This includes any downstream effects on the object from the interaction (e.g., gravity or inertia). For example, if we slide a block along a table, a valid $o_g$ is any block state before the block stops sliding. With an informative goal sampled, we can proceed to learn a goal-conditioned policy.}

\textbf{\REV{Extracting Affordances from Interaction for Robust Policy Learning}}: 
We define every interaction between the robot and the environment as exploiting some \textit{affordance} on an object in the scene. Affordances are properties of objects that define how they can be used (e.g., a block being grasped, a door knob being turned, or a drawer being opened).
\REV{Our insight is that learning these affordances (what happens to the object) instead of plans (what happens to the robot) from play will lead to a much simpler and more robust task representation that can operate over varying horizons, and thus will yield much better policies at test time. This paradigm empowers the policy to \textit{reason} about the environment: given access to an affordance (e.g., the door knob being turned) and the goal (e.g, opened door), the policy should be able to work backwards to infer the behavior to exploit that affordance (e.g., reach the knob and rotate the gripper to turn it). This is in contrast to prior work that relies on randomly selected, short, fixed horizon windows to learn latent representations of \emph{plans}---such plans fail to capture varying horizon tasks and overly depend on the robot state, leading to generalization issues at test time~\cite{lynch2020learning}.
}

\REV{\textbf{PLATO Design}: The PLATO architecture is shown in Fig.~\ref{fig:clfp}. At a high level, PLATO learns to model each interaction trajectory $\tau^{(i)}$ as a latent affordance $z$ (\textbf{(1) Posterior} in Fig.~\ref{fig:clfp}). This latent affordance $z$ \textit{explains} the actions both leading up to (pre-interaction) and during the interaction (see causal structure in Fig.~\ref{fig:clfp}). For example, knowing that we want to push a block (goal) and how we want to push it (affordance) allows the policy to infer that it should servo to the correct side of the block first. Therefore, the latent space is learned end-to-end with the policy $\pi$, which decodes latent affordances conditioned on the current state and object goal states to reproduce both pre-interaction and interaction robot actions (\textbf{(3) Policy on Interaction} and \textbf{(4) Policy on Pre-Interaction} in Fig.~\ref{fig:clfp}). By not learning to encode pre-interaction sequences, the critical assumption we make here is that variations in the affordances (interaction phase) are more relevant to the task and thus more important to capture in our latent space than variations across any random behavior sequence (pre-interaction).}

\REV{The policy action-reconstruction objective alone would encourage $z$ to model robot behaviors (plans), leading to a myopic view of the task. In order to force the latent space to focus on object affordances and thus be more robust at test time, we simultaneously learn a \textit{prior} on the affordance distribution conditioned on just the start and goal object states (\textbf{(2) Prior} in Fig.~\ref{fig:clfp}), trained to regularize the posterior latent affordance $z$. The posterior on $z$ considers the full window $\tau^{(i)}$, while the prior sees just the start and goal object states, and so the prior helps shape the latent space to encode the affordance in $\tau^{(i)}$ rather than just the action information.}

\vspace{-0.1cm} 
\begin{algorithm}
    \small
    \setstretch{1.35}
    \begin{algorithmic}[1]
    \floatname{algorithm}{Procedure}
    \STATE{Given: $H^{(i)}$, $H^{(-)}$, play data $D_{\text{play}}$, interaction criteria $f^{(i)}$,}
    \STATE{$D_{\text{play}}^{(-)},D_{\text{play}}^{(i)},D_{\text{play}}^{(+)} = f^{(i)}(D_{\text{play}})$ \hfill $\triangleright$ Split into interactions}
    \STATE{Initialize $E$, $E'$, $\pi$}
    \WHILE{not converged}
        \STATE{$\tau^{(-)}, \tau^{(i)}, \tau^{(+)} \sim D_{\text{play}}^{(-)},D_{\text{play}}^{(i)},D_{\text{play}}^{(+)}$}
        \label{alg:plato_main:line:sample}
        \STATE{Sample $o_g \sim \{o_t^{(+)}\}$}
        \label{alg:plato_main:line:sample_goal}
        \STATE{$p(z) \leftarrow E(\tau^{(i)})$  \hfill $\triangleright$ Posterior Affordance Distribution} \label{alg:plato_main:line:posterior}
        \STATE{$p(z') \leftarrow E'(o^{(i)}_1, o_g)$  \hfill $\triangleright$ Prior Affordance Distribution} \label{alg:plato_main:line:prior}
        \STATE{$z \sim p(z)$}
        \STATE{$\tilde{a}^{(i)}_{1:H^{(i)}} \leftarrow \pi(s^{(i)}_{1:H^{(i)}}, o_g, z)$  \hfill $\triangleright$ Policy in Interaction} \label{alg:plato_main:line:pol}
        \STATE{$\tilde{a}^{(-)}_{1:H^{(-)}} \leftarrow \pi(s^{(-)}_{1:H^{(-)}}, o_g, z)$  \hfill $\triangleright$ Policy in Pre-Interaction} \label{alg:plato_main:line:prepol}
        \STATE{Compute $\mathcal{L}_{\text{PLATO}}$ with Eq.~\eqref{eq:loss_plato} and update $\pi, E, E'$.}
    \ENDWHILE
    \end{algorithmic}
    \caption{PLATO Training}
    \label{alg:plato_main}
\end{algorithm}
\vspace{-0.2cm} 
\REV{The training procedure is outlined in Alg.~\ref{alg:plato_main}. After sampling windows from each interaction phase, $\tau^{(-)}$, $\tau^{(i)}$, and $\tau^{(+)}$ (Line~\ref{alg:plato_main:line:sample}), as well as a long term goal $o_g$ (Line~\ref{alg:plato_main:line:sample_goal}), PLATO encodes the interaction into a posterior and prior affordance distribution (Lines~\ref{alg:plato_main:line:posterior}-\ref{alg:plato_main:line:prior}). Next an affordance $z$ is sampled from the posterior using the reparameterization trick, and $z$ is then used to decode actions during interaction (Line~\ref{alg:plato_main:line:pol}) and pre-interaction (Line~\ref{alg:plato_main:line:prepol}). See Appendix~\ref{app:implementation:training} for more discussion of this training procedure, including its computational efficiency.} During training, we utilize action reconstruction losses over both the pre-interaction region and the interaction region windows to train all networks end-to-end. We utilize a KL divergence term to both train the affordance prior network and regularize the affordance posterior network.
\begin{align}
    \mathcal{L}_{\text{PLATO}} = -{\log(\pi(a^{(i)}_{1:H} | s^{(i)}_{1:H}, o_g, z))}
    -\alpha\ {\log(\pi(a^{(-)}_{1:H} | s^{(-)}_{1:H}, o_g, z))}
    + \beta\ \text{KL}(p(z) \parallel p(z')) \label{eq:loss_plato}
\end{align}
\REV{Here, $\alpha$ controls the policy focus on reconstructing pre-interaction behaviors, which usually is set to $1$. This procedure of learning a posterior and prior network and conditioning the policy on the latent context resembles existing work~\cite{lynch2020learning}. The key differences are i) our goals are sampled based on interactions with objects, ii) our latent representations capture object-centric affordances $z$ by intelligently shaping the learning objective, which can lead to more effective and generalizable policies, and iii) the policy reasons over varying and longer horizon sequences by conditioning on the same future desired affordance $z$ and long term goal $o_g$ throughout both pre-interaction and interaction.}

\noindent \textbf{Test Time:} At test time, the affordance posterior network $E$ cannot be used to output plans $z$, since it assumes access to a trajectory. Similar to prior work~\cite{lynch2020learning}, the prior network $E'$ is used to propose $z'$ at test time, using only the current object state $o_t$ and the desired goal object state $(o_g)$. Unlike prior work, the prior network only depends on object states, and will be robust to different robot starting states. Therefore instead of having to predict the exact robot behavior (i.e., plan) to achieve a goal, PLATO predicts an affordance at test time and empowers the policy to exploit this affordance. The policy conditions on $z'$ and goal $o_g$ to produce actions at the current state. 

\smallskip \noindent \textbf{Addressing Challenges of Prior Work}: By choosing goals $o_g$ resulting from interactions with objects, PLATO better reflects the true demonstrator goal, and thus can reduce the credit assignment problem found with fixed-horizon methods. PLATO handles variable horizon sequences by explicitly training the policy on pre-interaction and interaction sequences together, conditioned on the affordance and long term goal. By biasing the latent space to model \textit{affordances} during interaction, rather than just any random sequence of play, PLATO learns the task relevant behaviors and variations therein to aid in generalization. Our results in Sec.~\ref{sec:exp} demonstrate the effectiveness of PLATO compared to prior work on a wide range of complex tasks. See Appendix~\ref{app:challenges} for further discussion.

\section{Experiments}
\label{sec:exp}

\begin{figure}
    \centering
    \includegraphics[width=\linewidth]{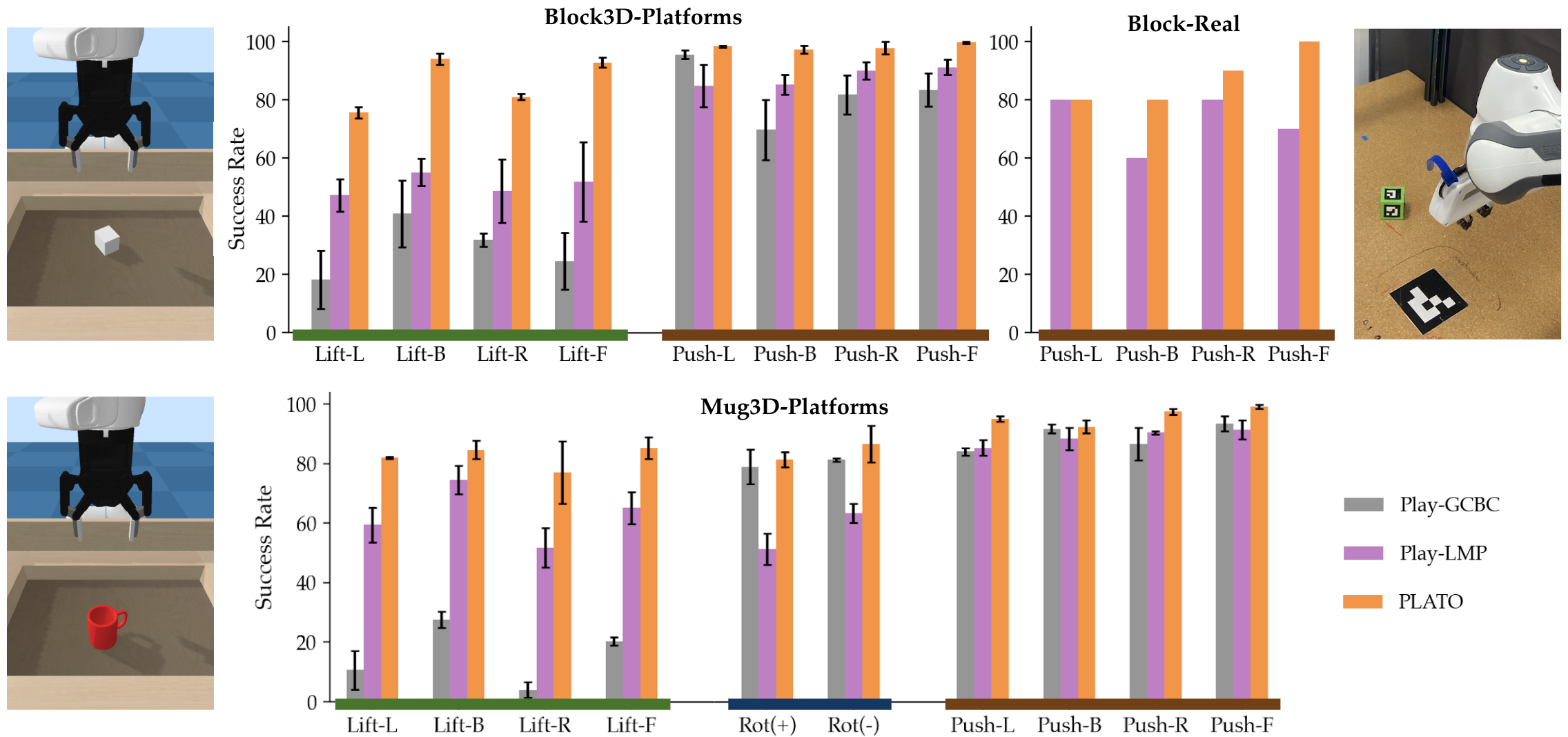}
    \caption{3D Environment Success Rates, trained over 3 random seeds and evaluated on various primitives for PLATO and baselines Play-LMP and Play-GCBC. \textbf{Top Left:} Block3D-Platforms. \textbf{Bottom:} Mug3D-Platforms. PLATO substantially outperforms baselines in 3D manipulation environments for pushing, rotating, and lifting tasks. \textbf{Top Right:} Block-Real. PLATO trained only in simulation generalizes to real world pushing tasks.}
    \label{fig:block3d_barplots}
    \vspace{-0.1cm}
\end{figure}

In this section, we evaluate our approach extensively across three single-object manipulation environments (\textbf{Block2D}, \textbf{Block3D-Platforms}, \textbf{Mug3D-Platforms}), one multi-object scene (\textbf{Playroom3D}), and one real scene (\textbf{Block-Real}), across diverse tasks like pushing, lifting, and rotating. These environments, shown in Fig.~\ref{fig:block3d_barplots}, enable a wide variety of objects and possible affordances. We collect scripted play data in all environments as well as human play data for \textbf{Block2D}, and train each method until convergence. See Appendix~\ref{app:experiments} for environment, task, data collection, and training details, and Appendix~\ref{app:implementation:arch} for method implementation details. 

\begin{figure}
    \centering
    \minipage{0.591\linewidth}
        \centering
        \includegraphics[width=.98\linewidth, trim={30pt 5pt 60pt 20pt}, clip]{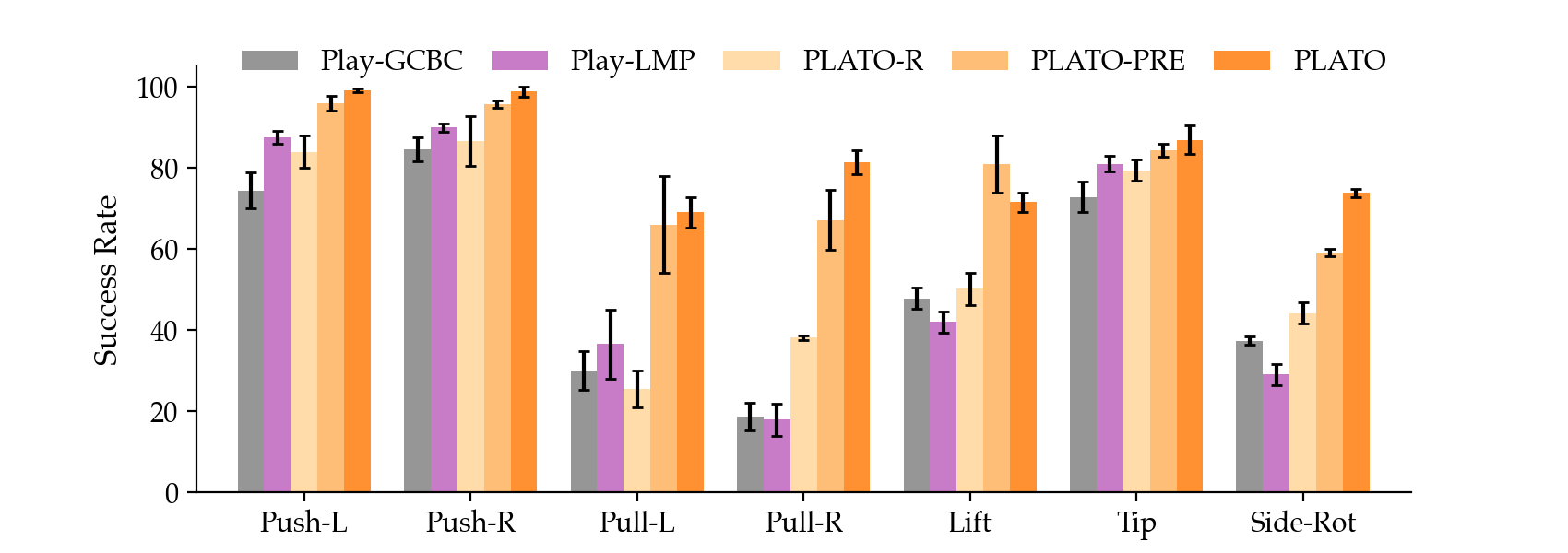}
    \endminipage%
    \hfill
    \minipage{0.409\linewidth}
        \centering
        \includegraphics[width=.99\linewidth, trim={35pt 5pt 20pt 20pt}, clip]{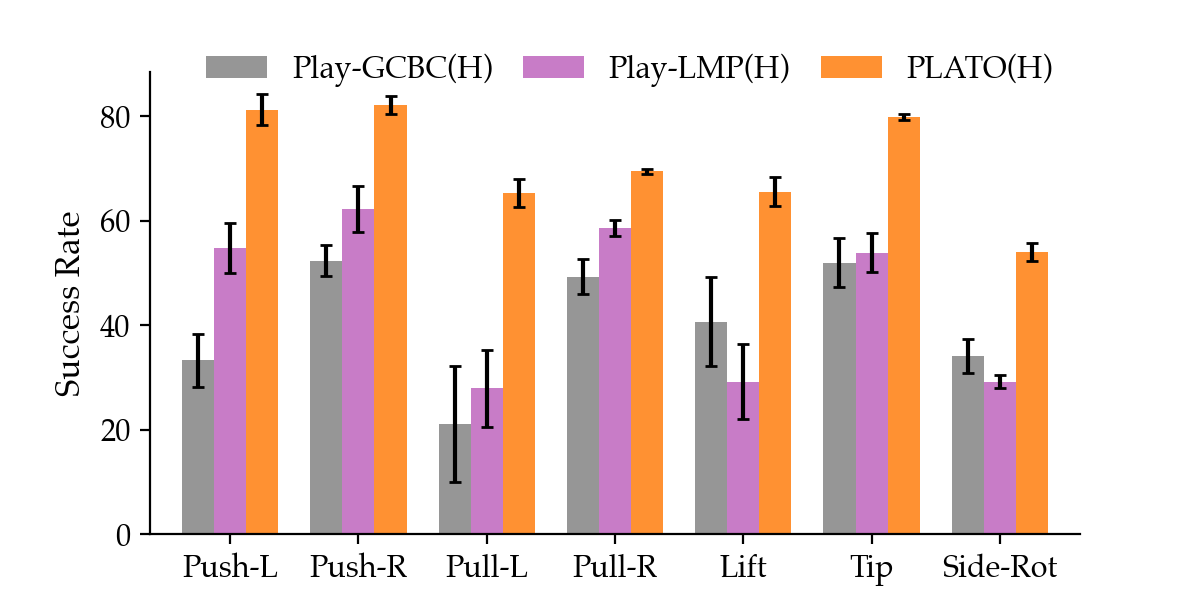}
        \captionsetup{width=.98\linewidth}
    \endminipage
    \caption{Block2D Success Rates, trained over 3 random seeds and evaluated on various primitives for PLATO and baselines Play-LMP and Play-GCBC. \textbf{Left:} Scripted play data, with ablations PLATO-PRE and PLATO-R.  \textbf{Right:} Human play data. PLATO substantially outperforms baselines on both scripted and human play data.}
    \label{fig:block2d_barplots}
    \vspace{-0.1cm} 
\end{figure}

\smallskip\noindent\textbf{Baselines:}
We compare PLATO against the two state-of-the-art methods for Learning from Play, Play-GCBC (Goal-Conditioned BC) and Play-LMP (Latent Motor Plans)~\cite{lynch2020learning}. 
We also implement two variants of our method, PLATO-PRE, which encodes both the interaction \textit{and} the pre-interaction periods into the latent space (adding $\tau^{(-)}$ as an input to $E$ in step (1) in Fig.~\ref{fig:clfp}), and PLATO-R, which replaces the object-centric prior with the prior from Play-LMP (adding the initial robot state as input to $E'$ in Fig.~\ref{fig:clfp}). PLATO-PRE results show the sufficiency of affordances as a latent representation, while PLATO-R results show how object-centrism benefits the learned latent space and the policy.


\smallskip \noindent \textbf{Block2D Results}: Evaluation results for PLATO for both scripted play data and human play data are shown in Fig.~\ref{fig:block2d_barplots}, along with results on PLATO-PRE and PLATO-R. On scripted play data, PLATO is able to substantially outperform the baselines on every task. Learning affordances from interactions enables PLATO to model much more complex action sequences, such as those involving the tether action, with high accuracy, even across variations in object dimensions, masses, and initial conditions.
On human play data, we again find that PLATO outperforms Play-GCBC and Play-LMP on all of the tasks, emphasizing the ability of our method to scale to human generated data. Interestingly, performance for all methods is worse on the human generated data than scripted data. We attribute this to human data containing many sub-optimal trajectories due to the challenges of teleoperation.

PLATO-PRE, which encodes the pre-interaction period, performs slightly worse but similar to PLATO, validating our hypothesis that interaction trajectories (PLATO) contain sufficient information about the task when compared to also including pre-interaction trajectories  in the latent space (PLATO-PRE).
We find that by adding robot state information to the prior (PLATO-R), performance suffers on most tasks. We hypothesize that this phenomenon is caused by the prior relying too much on the initial state of the robot, and not enough on that of the object. As a result, the latent space will not generalize at test time when the agent inevitably diverges from the offline state distribution.
Overall, we see that framing play data through the lens of object interactions (both PLATO and PLATO-PRE) results in much better policy learning than prior state-of-the-art methods.

\smallskip \noindent \textbf{Block3D-Platform \& Mug3DPlatform Results}: Fig.~\ref{fig:block3d_barplots} shows the results of evaluating PLATO on the harder Platform tasks. Interestingly, Both Play-GCBC and Play-LMP do well on the pushing tasks in this setting, but do very poorly on the more complex lifting tasks (and rotate tasks for Mug3D). By modeling an affordance space and properly relating these affordances to goal environment states, our method is capable of recreating diverse types of varying horizon behaviors from play, and unlike Play-LMP, our method scales smoothly as the number of tasks increases.

\begin{figure}
    \centering
    \includegraphics[width=\linewidth]{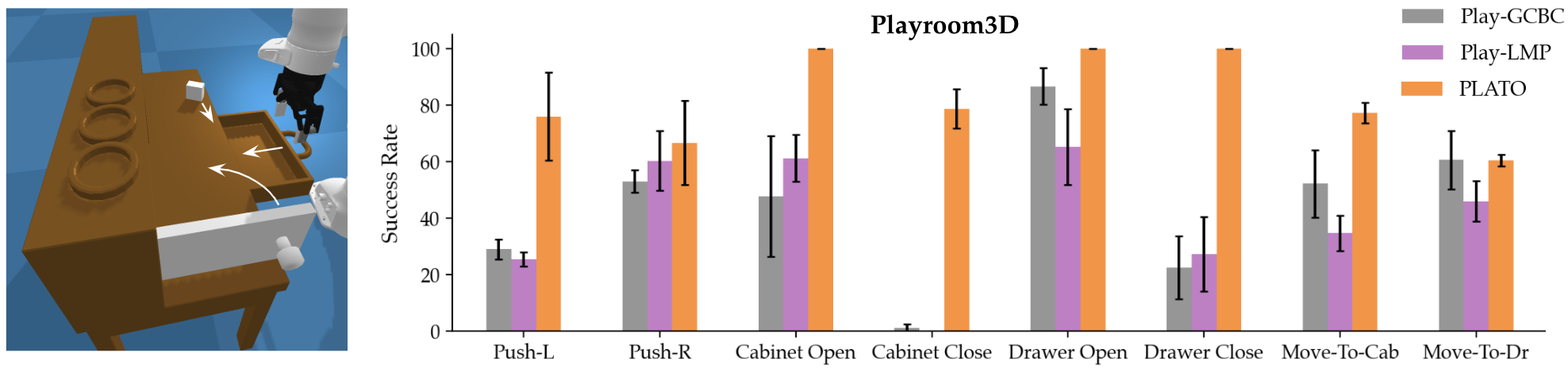}
    \caption{Playroom3D 3D Environment Success Rates. The difference between PLATO and baselines is even greater for more complex and varying horizon tasks, like opening and closing drawers and cabinets.}
    \label{fig:playroom3d_barplots}
    \vspace{-0.15cm}
\end{figure}

\smallskip \noindent \textbf{Playroom3D Results}: Fig.~\ref{fig:playroom3d_barplots} shows the results on the challenging Playroom3D environment with the drawer, cabinet, and the block, with especially diverse affordances and varying horizon tasks. Once again, PLATO outperforms the baselines on every task. While neither baseline can perform the complex Cabinet-Close primitive, PLATO achieves $100\%$ success. Notably, Play-GCBC performs better than Play-LMP on several tasks, suggesting that the prior network and policy might be out of distribution for the test time states and goals on these complex primitives.

\smallskip \noindent \textbf{Block-Real Results}:  Fig.~\ref{fig:block3d_barplots} shows the results of evaluating PLATO on pushing tasks for a real robot setup with \textit{no additional real world data} (see Appendix~\ref{app:experiments} for details). Play-LMP and PLATO both get near perfect success in simulation since these pushing tasks are less complex, but PLATO generalizes better across the gap between real and simulated object dynamics.

Overall, PLATO achieves substantially higher success rates than the baselines on a wide variety of tasks and object properties in simulation and real environments, along with lower variance across random seeds. This trend is even more apparent when we go beyond simple pushing tasks and consider more complex object interactions such as opening and closing doors, drawers, cabinets, lifting, and others. Due to its object-centric view of play, PLATO is able to extract and exploit diverse types of affordances in the environment and can handle varying horizon tasks. See Appendix~\ref{app:long_results} for additional experiments and a longer discussion of results.


\section{Conclusion}
\label{sec:conclusion}

\noindent \textbf{Summary}: In this work, we introduced an object-centric paradigm for learning from play data involving segmenting play into a series of object interactions. Prior state-of-the-art methods for learning from play suffer from credit assignment issues that stem from the short and fixed horizon of sampled trajectories. 
Our method, PLATO, addresses these credit assignment issues by choosing goals that involve meaningful changes in object state. PLATO learns a latent affordance space to model these interactions and their variations, and simultaneously learns to predict these latent affordances from the goal object state. These latent affordances help to inform the robot behavior across varying horizon tasks. Through our extensive experiments in both 2D, 3D, and real world environments spanning a wide variety of object manipulation tasks, we show that PLATO substantially outperforms prior methods on both scripted and human play data.

\smallskip 
\noindent \textbf{Limitations and Future Work}:
Our work introduces a paradigm of learning from variable horizon object interactions, and our method achieves substantially better performance across a variety of \textit{single-object} manipulation tasks. In future work, we intend to expand our notion of interaction to encompass even \textit{multi-object} interactions in play (e.g., tool use). When using tools, the robot might have second and third order effects on the environment that we could model. We believe the interaction paradigm introduced in this work is an important first step to reasoning about these higher order effects. 
For both the single and multi-object settings, future work might leverage notions of action information density or find bottleneck states to automatically segment interactions. 

As shown in Section~\ref{sec:exp}, data collection methods for play greatly affect final policy performance. We primarily evaluate with scripted play, but we hope to collect large human play datasets in future work. Compared to scripted play, we hypothesize that human play consists of much more behavior variability, yielding significant plan and affordance variability for a given start and goal state. Isolating affordances and interactions helps manage this variability, and PLATO is still able to perform well on all the tasks as a result. However, a future direction would be to study how exactly human play data differs from machine generated play data in order to develop more robust methods.

To add, our real robot experiments demonstrate PLATO trained in simulation can generalize to real world dynamics for pushing tasks without any data from the real robot. We hope to collect play data directly on the robot in future work to learn even more complex tasks in the real world. \REV{Furthermore, our method makes use of object state information, which may not always be easily available in practice. However, we claim this method can readily handle images with several simple changes to the learned prior, described in Appendix~\ref{app:limits}.}

\REV{We present thorough discussions of these limitations as well as potential solutions} in Appendix~\ref{app:limits}.



\vfill
\pagebreak
\bibliography{references}

\appendix

\section{Prior Work: Summary and Challenges}
\label{app:challenges}

Goal-conditioned behavior cloning on play trajectories $\tau \sim D_{\text{play}}$ (Play-GCBC) is quite effective at learning a robust multi-task policy; however, since the policy is only conditioned on the goal, it fails to capture all the degrees of behavior variation that would achieve that goal \cite{lynch2020learning}. Play-LMP addresses this by introducing a latent plan to capture not just \textit{what} goal to reach, but also \textit{how} to reach it. At a high level, Play-LMP learns to represent sampled play trajectories $\tau \sim D_{\text{play}}$ of fixed horizon $H$ as plans in a latent space, denoted by vector $z$, using variational inference techniques. The method trains a behavior cloning agent to reproduce the humans actions $a_{1:H} \in \tau$ conditioned on the current plan $z$ and the hindsight-labelled goal state $o_H$.

At a high level, Play-LMP encodes random sequences of environment states and robot actions as latent ``plans," which then get passed to a policy that learns to decode these plans at each state into the corresponding human demonstrated action for that time step. They simultaneously learn to predict the latent plan from just the initial and final state in the environment, for use at test time. In practice, the sequences are uniformly sampled 1-2 second chunks from play. Our method, PLATO, samples \textit{interactions} and intelligently learns a latent space from them, enable longer and variable horizon views of sequences of play.


\smallskip
\noindent \textbf{Architecture}: 
To represent play trajectories $\tau \sim D_{\text{play}}$ as plans, Play-LMP uses a posterior encoder $E$ that maps the states $s_{1:H} \in \tau$ to a single latent plan distribution $z \sim \mathcal{N}(\mu_{z}, \sigma_{z}^2)$. To regularize the posterior, Play-LMP simultanously learns prior network $E'$ that takes in just the first state $s_1$ and the goal environment state $s^o_H \in S^o$ to produce the latent plan distribution $z' \sim \mathcal{N}(\mu_{z'}, \sigma_{z'}^2)$. To decode plans into actions at a given state, Play-LMP utilizes a policy $\pi$ that takes in a plan $z$, sampled from either the prior distribution (test time) or posterior distribution (train time), along with the current state $s_t$ to predict the action $a_t$ at that step. 

$E$, $E'$, and $\pi$ are recurrent networks that operate over the fixed time horizon $H$ and are learned end-to-end at training time. This architecture can be viewed as ``encoding" trajectories in the environment into a lower dimensional plan $z$, and then reconstructing these plans into actions with the policy. PLATO similarly learns posterior, prior, and policy networks, but uses different inputs to each network based on sampling interactions and more meaningful goal environment states, as seen in Figure~\ref{fig:clfp}.

\smallskip
\noindent \textbf{Training}: $E$, $E'$, and $\pi$ are trained end-to-end by minimizing the following loss, equivalent to maximizing the evidence-based lower bound (ELBO) of the data likelihood under the posterior network $E$, given the learned prior network $E'$:
\begin{align}
    \mathcal{L}_{\textit{LMP}} = &-\frac{1}{H}\sum_{t=0}^{H-1}{\log(\pi(a_t | s_t, s^o_H, z))}
     + \beta\ \text{KL}(\mathcal{N}(\mu_{z}, \sigma_{z}^2) \parallel \mathcal{N}(\mu_{z'}, \sigma_{z'}^2))
\end{align}

There may be a variety of ways to go from the initial state $(s_0)$ to the final state $(s^o_H)$; for example, in order to move a block to the right, we might grab-move the block or push block without grabbing. Play-LMP benefits from encoding these variations in the latent space. PLATO uses a similar variational loss as Play-LMP, but instead of reconstructing actions from the short, fixed horizon $\tau$, reconstructs the interaction and pre-interaction trajectories $\tau^{(-}$ and $\tau^{(i)}$ conditioned on the affordance extracted from $\tau^{(i)}$ (see Eq.~\ref{eq:loss_plato}).

\smallskip
\noindent \textbf{Test Time}: At test time, the plan posterior network $E$ cannot be used to output plans $z$, since it assumes access to the full trajectory. Therefore, the plan prior network $E'$ is used to propose $z'$ at test time, using only the initial (current) state $s_t$ and the desired goal environment state $(s^o_H)$. The policy then uses the resulting distribution over $z'$ to predict actions online. 

\subsection{Play-LMP Strengths}

Play-LMP is able to interpret the effects of short interactions with the environment. In doing so, it learns to propose plan distributions $p(z')$ for achieving a wide variety of goals using just the learned prior $E'$. By capturing all the possible variations of going between each pair of start states $s_0$ and goal environment states $s^o_H$ during training, the Play-LMP policy is capable of executing a wide variety of behaviors at test time. As compared to Play-GCBC on sampled play windows, Play-LMP performs better on a wide range of manipulation tasks. Play-LMP and Play-GCBC are also shown be robust to minor perturbations in the environment, which can be attributed to the broad state-action-goal distributions found in human play data.

\subsection{Play-LMP Limitations}

\begin{figure}
    \centering
    \includegraphics[width=\linewidth]{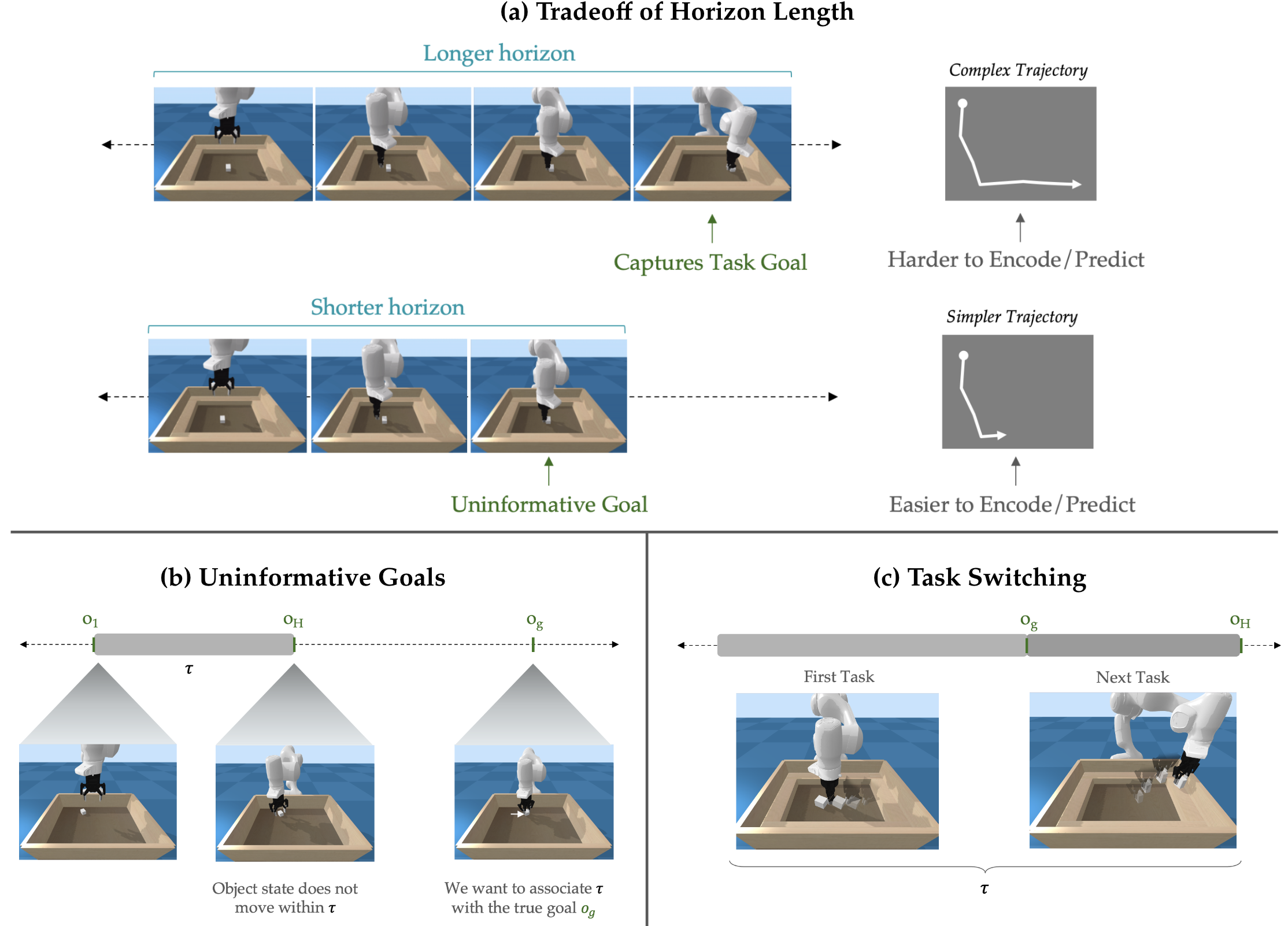}
    \caption{Challenges of short, fixed horizons, visualized for an example pushing task. (a) Choosing the horizon act is a balancing act between capturing task and goal information in the plan, and being able to predict plans (longer horizon means more complexity to capture) from just the initial and goal states. (b) If the horizon length is shorter than the current task, the goal environment state might be the same as the current state (if object state doesn't change within tau), and thus we will not properly link behaviors with the goals that induced these behaviors. (c) If the horizon is too long, We might capture multiple tasks within tau, and thus improperly assign the goal for the first task as being that of the next task. Our method, PLATO, addresses these challenges by biasing the latent space to learn from \textit{object interactions} within play.}
    \label{fig:challenges}
\end{figure}

Several conceptual issues arise from the goal labeling strategy. Play-LMP labels the last state in the sampled play sequence as the goal for all earlier states (hindsight relabelling). The assumption embedded in this method is that the last state of any sequence of length $H$ in play adequately represents the human's goal when choosing their actions (i.e., that the goal implies the actions). This can lead to the following issues relating to proper credit assignment:

\smallskip \noindent \textbf{Challenges of Fixed and Short Horizons: Inability to Capture Skills with Variable or Long Horizons}: As discussed, choosing the horizon length is a balancing act between picking capturing longer horizon task goals and behaviors in the latent space (posterior) and ensuring the predictability of those skills from just the start and goal states at test time (prior).
Due to this trade-off, Play-LMP can fail to capture the true goals when play consists of tasks whose horizon length is variable and/or large. In these settings, the longer horizon goal states will be out of distribution for the prior, and thus the policy can fail to produce the correct behavior. This trade-off in horizon lengths is visualized in Fig.~\ref{fig:challenges} (a) for a pushing example.

\smallskip \noindent \textbf{Challenges of Short Horizons: Uninformative Goals}: On the flip side, short sampled windows will often not contain any changes to the environment. One can imagine if the robot is re-orienting or servoing to an object, the goal state from the end of the short window will not be any different than the starting state, and thus uninformative for the prior network, even if the trajectory itself is nontrivial -- see Fig.~\ref{fig:challenges} (b), where $\tau$ only captures the reaching phase of pushing. Thus Play-LMP might learn to map many sequences in play to the null prior, $E'(s, s^o)$, where $s^o$ are the same environment state features in $s$.

\smallskip \noindent \textbf{Challenges of Fixed Horizons: Task Switching}: Another issue related to sampling windows randomly from play is that we might sample a window that contains a logical boundary between two tasks. Thus, the labelled goal (last state in the window) belongs to the second task, and will not necessarily inform the behavior in the first task -- see Fig.~\ref{fig:challenges} (c), where $\tau$ contains both a pushing and lifting task back to back. Thus the prior will likely not be able to predict the correct plan distribution for the policy to use, and will incur a high KL penalty that might shape the latent space disproportionately.

\smallskip \noindent \textbf{Challenges of Random Trajectory Sampling}: Even if we could perfectly label goals for each play trajectory $\tau$, it might be the case that not all sub-sequences should be “equal” in terms of their contribution to the latent space structure. For example, a sequence involving a robot reaching an object does not contain as many critical states -- states where the policy must be precise and accurate -- as a sequence involving picking up a block or rotating it in-hand. With Play-LMP, both of these types of sequences would be weighted equally in the construction of the latent space. We posit that the latent space should attend more to critical states in play like object interactions; therefore, prioritizing encoding sequences from object interactions will enable a more useful and information rich latent plan space. 

\smallskip 
Fundamentally, the issues above relate to a failure of \textit{credit assignment}, stemming from the short and fixed length of the sampled play sequences: the inferred goal does not actually represent the demonstrator's true goal. Our method PLATO considers tasks with variable horizon by leveraging object interactions with the environment. In the next section, we outline the implementation details for PLATO. 

\section{PLATO Implementation Details}
\label{app:implementation}

Next we will discuss the training procedure and low-level implementation details for PLATO, involving further algorithm details, network architectures, and hyperparameter choices.

\subsection{Training Procedure}
\label{app:implementation:training}

\begin{algorithm}
    \small
    \setstretch{1.35}
    \begin{algorithmic}[1]
    \floatname{algorithm}{Procedure}
    \STATE{Given: $H^{(i)}$, $H^{(-)}$, play data $D_{\text{play}}$, interaction criteria $f^{(i)}$,}
    \STATE{$D_{\text{play}}^{(-)},D_{\text{play}}^{(i)},D_{\text{play}}^{(+)} = f^{(i)}(D_{\text{play}})$ \hfill $\triangleright$ Split into interactions}
    \STATE{Initialize $E$, $E'$, $\pi$}
    \WHILE{not converged}
        \STATE{$\tau^{(-)}, \tau^{(i)}, \tau^{(+)} \sim D_{\text{play}}^{(-)},D_{\text{play}}^{(i)},D_{\text{play}}^{(+)}$}
        \STATE{Sample $o_g \sim \{o_t^{(+)}\}$}
        \STATE{$p(z) \leftarrow E(\tau^{(i)})$  \hfill $\triangleright$ Posterior Affordance Distribution} \label{alg:plato:line:posterior}
        \STATE{$p(z') \leftarrow E'(o^{(i)}_1, o_g)$  \hfill $\triangleright$ Prior Affordance Distribution} \label{alg:plato:line:prior}
        \STATE{$z \sim p(z)$}
        \STATE{$\tilde{a}^{(i)}_{1:H^{(i)}} \leftarrow \pi(s^{(i)}_{1:H^{(i)}}, o_g, z)$  \hfill $\triangleright$ Policy in Interaction} \label{alg:plato:line:pol}
        \STATE{$\tilde{a}^{(-)}_{1:H^{(-)}} \leftarrow \pi(s^{(-)}_{1:H^{(-)}}, o_g, z)$  \hfill $\triangleright$ Policy in Pre-Interaction} \label{alg:plato:line:prepol}
        \STATE{Compute $\mathcal{L}_{\text{PLATO}}$ with Eq.~\eqref{eq:loss_plato} and update $\pi, E, E'$.}
    \ENDWHILE
    \end{algorithmic}
    \caption{PLATO Training}
    \label{alg:plato}
\end{algorithm}

\REV{We now outline the training procedure for PLATO in greater detail, with Algorithm~\ref{alg:plato_main} reproduced above in Algorithm~\ref{alg:plato} for convenience. First, we sample segmented pre-interaction, interaction, and post-interaction periods from the play dataset. We then sample fixed length windows $\tau^{(i)} = (s_1,a_1,...,s_{H^{(i)}})$ from the interaction period and $\tau^{(-)} = (s_1,a_1,...,s_{H^{(-)}})$ from the pre-interaction period (Line 5 in Alg.~\ref{alg:plato}). Note that the post-interaction period is just the next pre-interaction period, and thus is still sampled for the next interaction. In the pull example in Fig.~\ref{fig:clfp}, $\tau^{(i)}$ is the pulling motion, and $\tau^{(-)}$ is the reaching motion before pulling.
We sample fixed horizon snapshots of each period for computational efficiency; however, the duration between $\tau^{(-)}$ and $\tau^{(i)}$ can vary tremendously, and so we are still able to capture variable and long horizon chains of events despite only sampling fixed horizon windows within each period.}

\REV{As described in Sec.~\ref{sec:method} in the main text, the goal object state $o_g$ for this chain of events can be selected as any object state after the interaction period's last object state \smash{$o^{(i)}_{H^{(i)}}$} and before the end of the post-interaction period (Line 6 in Alg.~\ref{alg:plato}).
During the interaction to post-interaction range, we know that the object trajectory is determined only by the interaction window actions and obstacles in the scene. For example, if we grab a block and then launch it along the table, the post-interaction period will consist of the block sliding; any state along that slide directly results from grabbing and launching. Thus the goal $o_g$ can correctly be attributed to affordance $z$.}

\REV{Having sampled $\tau^{(-)}$, $\tau^{(i)}$, and $o_g$, PLATO learns a goal-conditioned policy to reproduce the actions in both $\tau^{(-)}$ and $\tau^{(i)}$. Our insight is that much of the diversity in task-relevant behavior is contained during the interaction period, so instead of encoding $\tau^{(-)}$ and $\tau^{(i)}$ separately, we only encode $\tau^{(i)}$ with posterior $E$ (Line 7 in Alg.~\ref{alg:plato}). Now, the policy during interaction is conditioned on the $z \sim E(\tau^{(i)})$. Importantly, the policy during pre-interaction also conditions only on $z$, representing the \textit{future} interaction (Line 10-11 in Alg.~\ref{alg:plato}).}



\subsection{\REV{Architecture}}
\label{app:implementation:arch}

We follow a similar implementation as Play-LMP for the posterior, prior, and policy networks, as described in~\citep{lynch2020learning}. The posterior, prior, and policy networks are implemented as a Bidirectional GRU-RNN, an MLP, and a Unidirectional GRU-RNN, respectively. Input trajectories to the posterior include both robot and object state information, but aligning with Play-LMP we leave out actions for the posterior input, as including actions empirically worsens performance. Actions are target positions and orientations of the robot, since these are flexible and intuitive enough for humans to operate. All action reconstruction losses use Mean Absolute Error (deterministic actions), since empirically we found little benefit to using probabilistic actions with a negative log-likelihood loss. Specific architecture choices for each environment and method are detailed in Table~\ref{tab:arch}, as determined by extensive hyperparameter sweeps. For PLATO, we set $H^{(i)} = H^{(-)} = H$, and pre-interaction reconstruction loss weight $\alpha = 1$. Included in this hyperparameter sweep are minor horizon variations as employed in Relay Policy Learning~\cite{gupta2019relay}, which we found not to benefit policy learning for our settings. 

PLATO additionally uses a ``soft-boundary length" parameter ($S$) to allow for some flexibility in what is considered the boundary of interaction and pre-interaction during sampling. If $c_s$ and $c_e$ are the true contact start and end indices, then $\tau^{(i)}$ is sampled between $c_s - S$ and $c_e + S$. Likewise, $\tau^{(-)}$ is sampled from between $0$ and  $c_s + S$. This creates overlap between the pre-interaction and interaction regions, which we found empirically is necessary such that the policy can be trained contiguously across the contact border. In practice, we set $S = H/2$, since this allows for full coverage of the contact border during sampling.

\setlength{\tabcolsep}{3pt}
\renewcommand{\arraystretch}{1.25}
\begin{table*}[h!]
    \begin{center}
        \begin{tabular}{c|c|ccccccc}
            \multicolumn{1}{c}{\textbf{Environment}} & \multicolumn{1}{c}{\textbf{Method}} & $\beta$ & $H$ (seconds) & $|z|$ & $\pi$-hidden & $E$-hidden & $E'$-width & \\
            \Xhline{2\arrayrulewidth}
            \multirow{6}{*}{\textbf{Block2D}} & Play-GCBC & N/A & 2 & N/A & 128 & N/A & N/A & \\
             & Play-LMP & 1e-3 & 2 & 16 & 64 & 128 & 128 \\
             & PLATO & 1e-3 & 2 & 16 & 64 & 128 & 128 \\
            \cline{2-8}
             & Play-GCBC (H) & N/A & 2 & N/A & 256 & N/A & N/A & \\
             & Play-LMP (H) & 1e-3 & 2 & 16 & 256 & 128 & 256 \\
             & PLATO (H) & 1e-3 & 2 & 16 & 256 & 128 & 256 \\
             \hline
            \multirow{3}{*}{\textbf{3D-Flat}} & Play-GCBC & N/A & 4 & N/A & 128 & N/A & N/A & \\
            & Play-LMP & 1e-3 & 4 & 64 & 128 & 128 & 256 \\
            & PLATO & 1e-4 & 4 & 64 & 128 & 128 & 256 \\
            \hline
            \multirow{3}{*}{\textbf{3D-Platforms}}  & Play-GCBC & N/A & 4 & N/A & 128 & N/A & N/A \\
            & Play-LMP & 1e-4 & 4 & 64 & 128 & 128 & 256 \\
            & PLATO & 1e-4 & 3 & 64 & 128 & 128 & 256 \\
            \hline
            \multirow{3}{*}{\textbf{Mug-3D}}  & Play-GCBC & N/A & 4 & N/A & 256 & N/A & N/A \\
            & Play-LMP & 1e-4 & 4 & 64 & 256 & 256 & 256 \\
            & PLATO & 1e-4 & 4 & 64 & 256 & 256 & 256 \\
            \hline
            \multirow{3}{*}{\textbf{Playroom3D}}  & Play-GCBC & N/A & 4 & N/A & 256 & N/A & N/A \\
            & Play-LMP & 1e-3 & 4 & 64 & 256 & 256 & 256 \\
            & PLATO & 1e-4 & 4 & 64 & 256 & 256 & 256 \\
        \end{tabular}
        \caption{Best performing hyperparameters for each environment, method, and data source. $\beta$ controls the regularization on the posterior from the learned prior. Each method is quite sensitive to this amount of regularization. $H$ is the horizon length and controls the length of the sampled trajectory for the posterior encoder. All methods are quite sensitive to this as well. $z$ is the latent vector, and $|z|$ is the latent dimensionality. $\pi$-hidden and $E$-hidden are the hidden sizes for $\pi$ and $E$ respectively, and these control the expressiveness of each network. $E'$-width is the width of the prior network. We do not vary the number of layers in each network, which are chosen to be the same as in prior work.}
        \label{tab:arch}
    \end{center}
\end{table*}

\section{Experimental Details}
\label{app:experiments}

In this section we outline the environments, our real world setup, tasks, data collection, and evaluation process we employed. Each environment has substantial variability, and scripted policies are similarly designed to be quite diverse with sizeable injected noise.

\begin{figure*}[h!]
    \centering
    \includegraphics[width=\linewidth]{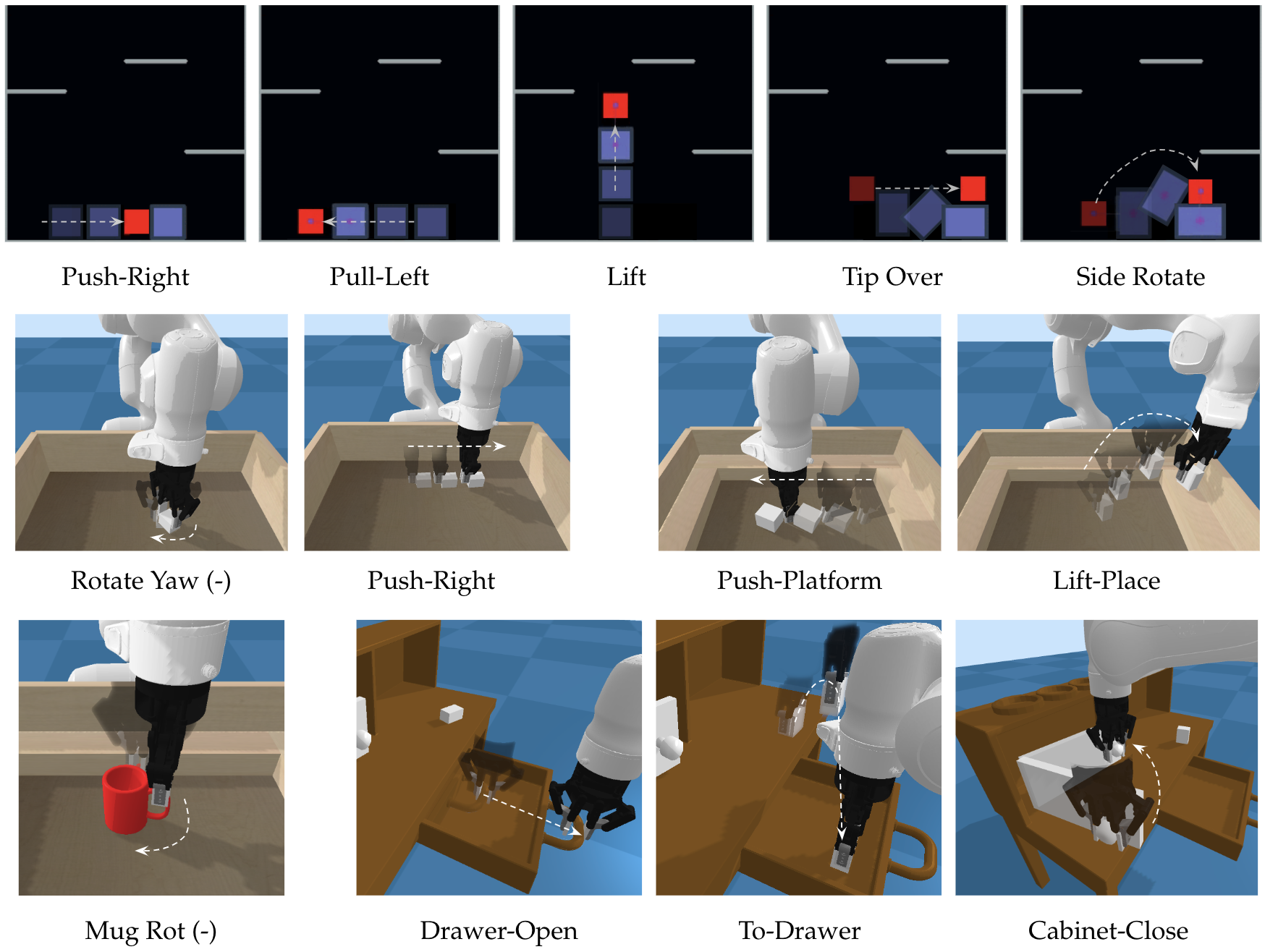}
    
    \caption{\underline{\textbf{First Row}}: Block2D Environment Primitive Examples. Each of these primitives can be executed from a variety of object initial conditions, masses, and dimensions. \underline{\textbf{Second Row}}: Block3D and Block3D-Platform Primitive Examples. Again, object initial conditions, masses, and dimensions are varied during play. The left two primitives shown are taken from \textbf{Block3D-Flat}, and the right two are taken from \textbf{Block3D-Platforms}. These represent a subset of the evaluation primitives in the 3D environment, and are meant to show the diversity of tasks and behaviors our method is evaluated on. \underline{\textbf{Third Row}}: The left image shows an example primitive in \textbf{Mug3D-Platforms}. The right three images show sample tasks from \textbf{Playroom3D}.}
    \label{fig:primitives}
    
\end{figure*}

\subsection{Environments and Tasks}

\REV{For all simulated environments, the contact signal used in simulation is the binary contact information between the robot and the rest of the scene, which can easily be computed in pybullet (3D) and pymunk (2D).}

\smallskip \noindent \textbf{Block2D}: The first environment is a 2D continuous block manipulation environment implemented with the PyMunk 2D physics engine~\cite{pymunk}. Blocks in the environment are sized, massed, and positioned randomly at every reset. The ego agent (red) can interact with these blocks in the presence of gravity, with a special ``tether" action that creates a link constraint if the ego agent is close enough to the block. This environment is meant as a 2D analog to more challenging block manipulations. As shown in Figure~\ref{fig:primitives}, we constructed a set of block manipulation primitives to evaluate on: Push, Pull, Lift, Tip, and Side-Rotate.

\smallskip \noindent \textbf{Block3D-Flat}: The second environment is a 3D block manipulation environment, involving a simulated Franka Emika Panda robot arm, 3D blocks, and a table playground area surrounded by walls. Similarly to Block2D, the block dimensions, initial positions, and masses here are randomly sampled. The Panda robot arm uses an operational space torque controller to exert realistic and bounded forces on the blocks. In this environment, we implement two dimensional Push primitives along the table surface, as well as a Top-Rotate primitive to control the z-axis rotation of the block in either direction (see Figure~\ref{fig:primitives}). Results for this environment, shown in Table~\ref{tab:block3dflat}, were not included in the main text due to space limitations.

\smallskip \noindent \textbf{Block3D-Platforms}: The third environment builds on Block3D-Flat, but introduces platforms along the walls to enable even more complex primitives like lifting and placing. Here, we implement two dimensional Push primitives, and Lift-Place primitives in all cardinal directions on the table plane (see Figure~\ref{fig:primitives}).

\smallskip \noindent \textbf{Mug3D-Platforms}: The fourth environment is like the Block3D-Platforms environment, but uses a challenging mug object of varied size and mass, instead of blocks. Here, we implement two dimensional Push primitives, and Lift-Place primitives in all cardinal directions on the table plane, and additionally a Mug Rotate primitive similar to Block3D-Flat (see Figure~\ref{fig:primitives} for an example of Mug Rotate). These primitives each require unique types of affordances for a mug, and this environment is designed to test how learning from play methods can adapt to more precise manipulation tasks.

\smallskip \noindent \textbf{Playroom3D}: The fifth environment is similar to the one used in prior work, involving several dynamic objects: a block on the table, a drawer, and a cabinet door. This environment is challenging since it involves learning affordances over multiple objects, as well as interactions between them (like putting objects into the drawer or cabinet). For the cabinet, we test opening and closing actions, and likewise for the drawer. For the object, we test Pushing Left/Right primitives as well as moving the object To-drawer and To-cabinet  (see Figure~\ref{fig:primitives} for examples). These actions require many time-steps and have several bottleneck states. Play data also consists of Take-out-drawer and Take-out-cabinet, but we are not currently evaluating on these.

\begin{figure}
    \centering
    \includegraphics[width=0.7\linewidth]{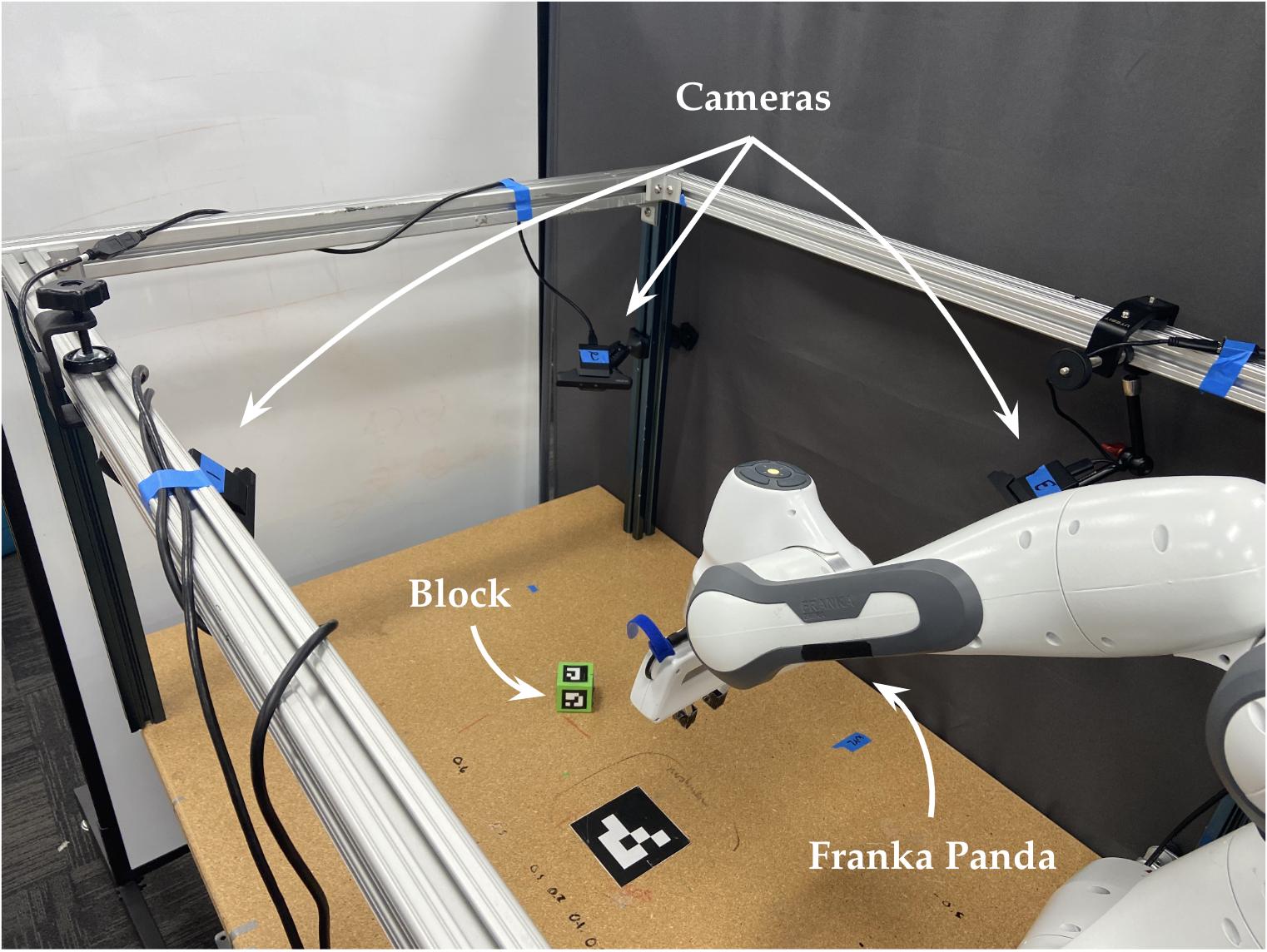}
    \caption{Block-Real environment. We use the Franka Emika Panda 7DOF robot arm for our experiments. The green cube we use for block manipulation tasks is shown in the middle, with ArUco tags on each face for pose detection. The 6D pose of the object is estimated via a multi-camera setup in order to be robust to occlusion by the robot, as shown here with the camera on the far right. The ArUco tag in the middle is used to calibrate the extrinsics of the cameras and localize the object frame of reference relative to the robot.}
    \label{fig:rw_setup}
\end{figure}

\smallskip \noindent \textbf{Real Robot Environment}: Finally, we create a real robot environment that mirrors our simulation environment. Our setup is shown in Figure~\ref{fig:rw_setup}. We use three mounted Realsense SR300 cameras to robustly detect the pose of the green cube in the presence of occlusions from the robot, using OpenCV's Aruco tag detection for 3D pose estimation.  As with our simulation experiments, we condition policies on full state of the environment (object 6D pose + end effector 6D pose + gripper state) rather than images. We add noise to the object state in simulation to match the noise seen in real world block state estimation. This along with an identical action space helps to reduce the sim-to-real gap and enables immediate deployment of simulation trained policies in the real world. Regardless, the dynamics of the object are still starkly different than those in our simulation dataset: the block is lightweight and slightly deformable, and has high friction with the table that can cause it to flip over instead of slide on the table. We test pushing primitives in this environment to demonstrate that PLATO is scalable to real world tasks with minimal data augmentation.

\REV{\textbf{Real-World Deployment}: To deploy this system more generically (including training on real world data), there are several key infrastructure additions beyond the setup we have shown here. In terms of hardware, the robot would need an additional contact sensing patch on the gripper or a force/torque sensor at the end effector to automatically detect interaction with the scene, as well as several cameras to observe the scene. Note that contact readings are only used during training (for the purpose of interaction segmentation). Since our real world setup did not involve training on real world data, we did not need to add these additional sensors. However, we believe this modification should be quite straightforward. With pressure sensing (which is closest to what we do in simulation), there is a limitation that only interactions with the pressure sensing portion of the end effector will be counted during segmentation. In terms of software, a robust object 6D pose detection algorithm would be utilized to detect the object states under potential occlusion. With these additions, our method should readily scale to many real robot systems. }

Example Task primitives for simulation evaluations are shown in Fig.~\ref{fig:primitives}. There is substantial within-task noise for scripted data to more closely resemble human data and real world conditions.

\subsection{\REV{Task Complexity}}

\REV{We believe our set of tasks covers a wide spectrum of levels of complexity, from simpler pushing tasks to more complex door opening or object rotation tasks.}

\REV{Firstly, we have incorporated diversity in primitives and variations in objects, which we emphasize is crucial and is not present in prior work such as Play-LMP. The tasks themselves cover a number of diverse primitives, where each “primitive” consists of variations in the goal, for example varying pushing distance or lifting placement location. The motions of the scripted primitives also have sizable variations in intermediate waypoints, speed, etc. Within all of our environments, we inject large variations in the positions, orientations, each size dimension, and masses of each of the blocks and mugs, which each require different strategies from the robot’s perspective. Furthermore, grasping the mug involves a very precise interaction with the handle, which contrasts the wide, centered grasp used with blocks. This is in comparison to the closest prior work, i.e., Play-LMP tasks, which usually are projected to far fewer primitives and variations in the policy.  The Play-LMP tasks largely involve either fixed object shapes with limited random pose initialization or static scene elements like buttons and constrained unchanging drawers. This might make it seem that these tasks are complex at the surface visual level, but we argue that our set of tasks and primitives require a much greater range of behaviors, such as grasping different shapes, rotating blocks to a wide spectrum of new orientations, and handling a variety of block masses, and thus these tasks are more complex. We would like to emphasize that visually interesting environments (e.g., added buttons or static objects as in Play-LMP), do not really add to the complexity of policy learning. What makes policy learning challenging is variations in behaviors and object properties, which we extensively test with our experiments. We believe that the performance of Play-LMP suffers in these settings precisely because the tasks are more complex for policy learning.} 
 
\REV{Secondly, our experiments include a visually interesting and complex environment, Playroom3D, which represents a more challenging version of the tasks used in Play-LMP involving some of the same underlying assets but having multiple randomized objects. To make the tasks even more challenging, we added the cabinet door opening and closing tasks. This “door opening” affordance was not included in the Play-LMP prior work, and is significantly harder to learn (see Play-LMP and Play-GCBC performance). We thus believe that our tasks are significantly more complex and diverse compared to prior work in this domain and adequately demonstrate the performance of our algorithm and a significant gap with prior work.}

\subsection{Evaluation Details}
To evaluate each method, we evaluate across a set of environment-specific primitives by first running the primitive at the current state to generate a goal, then resetting and running the policy conditioned on that goal. \REV{We separately evaluate tasks by what we considered to be ``similar" affordances. For example, push-left and push-right are considered as two separate tasks. However, there is sizeable variation in the set of goals that comprise each individual task. For example, with pushing, there is variation in the pushing distance, as well as all inherent variations in object properties in the environment (e.g., mass, shape, etc) discussed previously. Figure~\ref{fig:tsne} provides some intuition that our affordance space learns to cluster by these directions}. Success metrics are specified according to the task primitive being tested, usually involving the distance to either the goal object position or orientation. Evaluation times out after a fixed number of steps if the given method is unsuccessful. For stability, the latent vector $z'$ is sampled once every $T \leq H$ steps and held constant during action decoding by the policy for the next $T$ steps. For Play-LMP and PLATO, latent vectors are recomputed at the same frequency during evaluation, but the goal is fixed for the evaluation period.

\subsection{Data Collection}
To evaluate our method across a wide variety of affordances available in the environment, we collect a large dataset of play data (roughly 15 hours) under artificial demonstrator agents that repetitively and randomly choose between the hand designed primitives for each environment. \REV{Proprioceptive state information includes robot end-effector 6D poses and twists, and object state information includes the object 6D poses, the 1D cabinet angle, and the 1D drawer open distance.} In order to replicate human play as closely as possible, we add large amounts of variation to the primitives through randomized parameters. For example, with the pulling primitive, we vary the speed and duration of the pulling motion, as well as the reaching behavior. The boundaries between primitives are assumed to be unknown during training, like in human play data. 

We also separately collect a smaller dataset (roughly 8 hours) of noisy human play data to evaluate the Block2D environments. From these experiments, we draw insights on the differences between human and scripted play data. \REV{Human play data for the Block2D task is collected using a keyboard control interface. Arrow keys control the ego agent position, while ‘g’ controls the grabbing tether action. One proficient user was used to collect the data, and was shown a set of tasks (the evaluation tasks) to perform during play with the instruction of trying to equally represent each task in their play, but they were also clearly informed that they were not limited to performing just these tasks. This user was given 15 minutes to practice in the environment, after which point data collection began. In future work, we hope to reduce the dependence on balanced, curated datasets and allow play to be truly freeform.}

\section{Results and Analysis}
\label{app:long_results}

\newcommand{\MS}[2]{#1(#2)}
\newcommand{\MSB}[2]{\textbf{#1(#2)}}

In this section we discuss in greater depth our results, additional experiments, and tables with exact success rates to complement the bar-plot figures in the main text (Figures~\ref{fig:block2d_barplots}, \ref{fig:block3d_barplots}, \ref{fig:playroom3d_barplots}). 

\subsection{\REV{Detailed Results}}
\label{app:detailed_results}

\renewcommand{\arraystretch}{1.25}
\setlength{\tabcolsep}{3pt}
\begin{table*}[h]
    \begin{center}
        \begin{tabular}{c|ccccccc}
            
            \multicolumn{1}{c}{} & \textbf{Push-L} & \textbf{Push-R} & \textbf{Pull-L} & \textbf{Pull-R} & \textbf{Lift} & \textbf{Tip} & \textbf{Side-Rot} \\
            \Xhline{2\arrayrulewidth}
            Play-GCBC & \MS{74.4}{4.4} & \MS{84.5}{2.9} & \MS{30.0}{4.8} & \MS{18.6}{3.4} & \MS{47.8}{2.6} & \MS{72.8}{3.8} & \MS{37.3}{1.0}  \\
            
            Play-LMP & \MS{87.5}{1.5} & \MS{89.9}{1.0} & \MS{36.5}{8.5} & \MS{17.9}{4.0} & \MS{42.0}{2.6} & \MS{81.0}{2.0} & \MS{29.0}{2.7} \\
            
            PLATO-R & \MS{83.9}{4.0} & \MS{86.6}{6.12} & \MS{25.4}{4.5} & \MS{38.1}{0.5} & \MS{50.2}{4.0} & \MS{79.4}{2.6} & \MS{44.2}{2.7}\\
            
            PLATO-PRE & \MS{95.9}{1.9} & \MS{95.7}{1.0} & \MS{66}{11.9} & \MS{67.1}{7.4} & \MSB{80.9}{7.1} & \MS{84.3}{1.6} & \MS{59}{0.9}  \\
            
            PLATO & \MSB{99.1}{0.5} & \MSB{98.8}{1.2} & \MSB{69.0}{3.8} & \MSB{81.4}{3.0} & \MS{71.5}{2.3} & \MSB{86.9}{3.6} & \MSB{73.8}{1.0}  \\
            \hline
            Play-GCBC (H) & \MS{33.3}{5.1} & \MS{52.4}{3.0} & \MS{21.1}{11.1} & \MS{49.3}{3.4} & \MS{40.7}{8.6} & \MS{52.0}{4.7} & \MS{34.1}{3.3} \\
            
            Play-LMP (H) & \MS{54.8}{4.7} & \MS{62.2}{4.4} & \MS{27.9}{7.4} & \MS{58.6}{1.6} & \MS{29.2}{7.2} & \MS{53.9}{3.7} & \MS{29.2}{1.2} \\
            
            PLATO (H) & \MSB{81.3}{3.0} & \MSB{82.2}{1.7} & \MSB{65.3}{2.7} & \MSB{69.5}{0.5} & \MSB{65.6}{2.8} & \MSB{79.9}{0.6} & \MSB{54.0}{1.7} \\
        \end{tabular}
        \caption{Block2D Success Rates in percentages in the form \MS{mean}{std-err}, trained over 3 random seeds and evaluated on various Push, Pull, Lift, Tip, and Side-Rotate primitives. Block sizes are randomized in each dimension, and blocks are initialized in random positions along the bottom of the grid. Our method PLATO outperforms all prior methods on both scripted and human data. PLATO-PRE includes pre-interaction information in the learned latent space (amounting to passing both $\tau^{(i)}$ \textit{and} $\tau^{(-)}$ into the posterior $E$), but increases the training time compared to PLATO. PLATO-R incorporates the current robot state into the prior, representing the non object-centric version of PLATO. Object-centric methods that learn from interactions (PLATO, PLATO-PRE) perform much better than their counterparts.}
        \label{tab:block2d}
    \end{center}
\end{table*}

\renewcommand{\arraystretch}{1.25}
\setlength{\tabcolsep}{3pt}
\begin{table*}[h]
    \begin{center}
        \begin{tabular}{c|ccccccc}
            
            \multicolumn{1}{c}{} & \textbf{Push-L} & \textbf{Push-R} & \textbf{Pull-L} & \textbf{Pull-R} & \textbf{Lift} & \textbf{Tip} & \textbf{Side-Rot} \\
            \hline
            PLATO & \MS{99.1}{0.5} & \MS{98.8}{1.2} & \MS{69.0}{3.8} & \MS{81.4}{3.0} & \MS{71.5}{2.3} & \MS{86.9}{3.6} & \MS{73.8}{1.0}  \\
            \hline
            \REV{PLATO-FC(4\%)} & 100 & 95.9 & 61.3 & 57.0 & 78.4 & 93.0 & 68.3 \\
            
            \REV{PLATO-FC(8\%)} & 97.0 & 96.7 & 35.0 & 65.6 & 80.4 & 95.6 & 48.7 \\
            
            \REV{PLATO-FC(12\%)} & 94.8 & 100 & 61.5 & 10.1 & 60.8 & 92.3 & 63.6 \\
        \end{tabular}
        \caption{\REV{Contact Signal Ablation Experiment. Here we artificially add fake contact signals outside of interactions (e.g., during pre-interaction), causing false interactions to be segmented during training. We evaluate PLATO-FC(\%), where \% denotes the percentage of \textit{interactions} that are false positives. We show results for a single seed of each PLATO for 4\%, 8\%, and 12\%. Pulling tasks have some variance in performance under added false contact, however considering how much data is affected, PLATO is quite robust for all tasks.}}
        \label{tab:block2d_ablate_contact}
    \end{center}
\end{table*}

\smallskip \noindent \textbf{Block2D}:
In Table~\ref{tab:block2d}, we see that PLATO substantially outperforms the baselines on each task. Play-LMP and Play-GCBC get roughly similar performance on most tasks, and struggle the most on tasks involving the tether action (Pulling and Side-Rotate). 
To further study the effects of encoding just the interaction period in the latent space, we implement PLATO-PRE, a version of PLATO in which the posterior encodes sampled trajectories from both the pre-interaction period $\tau^{(-)}$ and interaction period $\tau^{(i)}$, instead of just the interaction period. We see that PLATO and PLATO-PRE do roughly equivalently on average, with PLATO doing substantially better on Side-Rotate, but PLATO-PRE doing better on the lift primitive. Note that PLATO-PRE is slower to train since the posterior recurrent encoder receives a much longer sequence. We hypothesize that this is due to the tradeoff of various tasks between the complexity of the interaction and the complexity of pre-interaction behaviors. Overall, we can conclude that the including pre-interaction trajectories in the latent space (PLATO-PRE) is not uniformly better than only including interaction trajectories (PLATO), and thus does not warrant the added training time. This confirms our intuition that for complex interaction sequences like Side-Rotate, the interaction period contains enough information from the perspective of representation learning. Importantly, we see that framing play data through the lens of object interactions (PLATO, PLATO-PRE) results in much better policy learning than prior state-of-the-art methods.

Interestingly, performance for all methods is worse on the human generated data than scripted data. We attribute this to the fact that despite the significant noise added to the scripted primitives during data collection, scripted data still has cleaner and more successful demonstrations of each task than human data, which can contain many sub-optimal trajectories due to the challenges of teleoperation. For example, we observed that the Side-Rotate primitive is only successful in the human play dataset around 70\% of the time. Additionally, humans can demonstrate the same task in many more ways than we could possibly script (e.g., by elongating the duration of a pulling motion or picking a wildly different spot to pull from), resulting in much more complex plans and policies to learn. This is supported by the fact that, as shown in Table~\ref{tab:arch}, the best performing architectures for each method on human play data involved larger policy hidden sizes as compared to the best performing methods on scripted play data.

\begin{figure}
    \centering
    \includegraphics[width=0.9\linewidth]{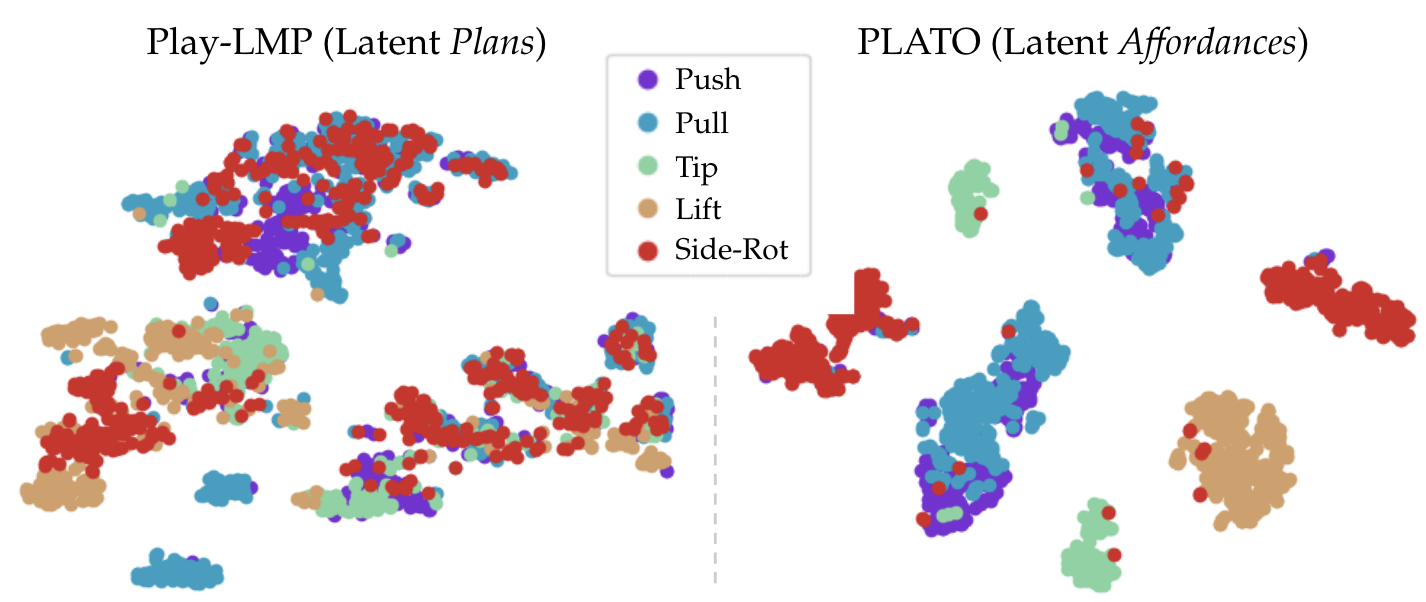}
    \caption{Learned latent spaces for Play-LMP (left) and our method PLATO (right) in the Block2D environment, plotted in 2 dimensions (from original 16 dimensions) with t-SNE. While there is overlap between tasks in the learned \textit{plan} space in Play-LMP, tasks are much more separable with the learned \textit{affordance} space in PLATO. We see separate clusters for each primitive likely relating to the direction of the primitive, and we also see within-cluster variation for each primitive. Note that push and pull have sizeable overlap since these represent similar motions (motion left and right on the ground). This clustering suggests that learning from interactions helps the posterior $E$ to provide simpler and more informative latent representations for the policy.}
    \label{fig:tsne}
\end{figure}

For these 2D environments, we additionally examine the structure of the latent space learned by PLATO compared to prior work. Figure~\ref{fig:tsne} shows example learned latent spaces for both Play-LMP and our method PLATO on Block2D; we see that by training on interactions, PLATO recovers a latent space that is more separable by task than Play-LMP despite not being presented with any task labels during training.

\REV{Additionally, in Table~\ref{tab:block2d_ablate_contact}), we ablate the quality of the interaction signal (contact) in order to determine the important of clean contact signals. To accomplish this, we add in \textit{false positive} contact signals to the data (e.g., during pre-interaction). This causes some percentage of the interactions sampled during training (denoted by FC(\%) in Table~\ref{tab:block2d_ablate_contact}) to actually be non-contact sequences. We see that PLATO is quite robust to added false positive contacts. The Pull task has high variance in performance here under added contacts, however there is no clear relationship between more contact noise and worse performance for any of the tasks. This is likely due to the fact that all models condition on the start and goal states, and false contact signals usually involve no change in the object state; therefore, from the model's perspective, sampled false interactions will be completely distinguishable from the task affordances that we care about at test time.}

\renewcommand{\arraystretch}{1.25}
\begin{table*}[h]
    \begin{center}
        \begin{tabular}{c|cccccc}
            \multicolumn{1}{c}{} & \textbf{Push-F} & \textbf{Push-L} & \textbf{Push-B} & \textbf{Push-R} & \textbf{Rotate (+)} & \textbf{Rotate (-)} \\
            \Xhline{2\arrayrulewidth}
            GCBC & \MS{71.3}{3.0} & \MS{65.7}{10.0} & \MS{57.5}{11.2} & \MS{50.1}{7.8} & \MS{53.7}{2.6} & \MS{42.4}{1.9} \\
            
            LMP & 
            \MS{68.0}{4.5} &
            \MS{63.0}{4.4} & \MS{50.6}{2.3} & \MS{28.0}{9.0} & \MS{39.6}{3.2} & \MS{40.7}{3.4} \\
            
            PLATO & \MSB{77.4}{4.3} & \MSB{89.4}{5.5} & \MSB{74.0}{4.7} & \MSB{84.0}{5.0} & \MSB{83.3}{4.4} & \MSB{78.6}{6.7} \\
        \end{tabular}
        \caption{Block3D-Flat Success Rates in percentages in the form \MS{mean}{std-err}, trained over 3 random seeds and evaluated on pushing and rotation tasks. Again, our method PLATO outperforms all prior methods, especially on the harder rotation primitives. Interestingly, we also see that Play-LMP does not do as well as Play-GCBC on several primitives, likely due to the prior and policy being out of distribution at test time.}
        \label{tab:block3dflat}
    \end{center}
\end{table*}

\smallskip \noindent \textbf{Block3D-Flat}: Results for this block pushing and rotating environment (not presented in the main text due to space limitations) are shown in Table~\ref{tab:block3dflat}. Here again, PLATO is able to significantly outperform the baselines on all of the tasks, especially in the more fine-grained rotation motions, as well as on the pushing tasks. 

\renewcommand{\arraystretch}{1.25}
\setlength{\tabcolsep}{2.9pt}
\begin{table*}[h]
    \begin{center}
        \begin{tabular}{c|cccccccc}
            \multicolumn{1}{c}{} & \textbf{Push-F} & \textbf{Push-L} & \textbf{Push-B} & \textbf{Push-R} & \textbf{Lift-F} & \textbf{Lift-L} & \textbf{Lift-B} & \textbf{Lift-R} \\
            \Xhline{2\arrayrulewidth}
            GCBC & \MS{95.5}{1.5} & \MS{69.7}{10.4} & \MS{81.8}{6.7} & \MS{83.4}{5.6} & \MS{18.2}{10} & \MS{40.8}{11.4} & \MS{31.9}{2.3} & \MS{24.6}{9.8} \\
            
            LMP & \MS{84.7}{7.3} & \MS{85.3}{3.4} & \MS{90}{2.9} & \MS{91.2}{2.6} & \MS{47.2}{5.6} & \MS{55.1}{4.6} & \MS{48.6}{10.9} & \MS{51.8}{13.6} \\
            
            PLATO & \textbf{\MS{98.3}{0.4}} & \textbf{\MS{97.2}{1.4}} & \textbf{\MS{97.8}{2.2}} & \textbf{\MS{99.7}{0.3}} & \textbf{\MS{75.6}{1.9}} & \textbf{\MS{94.0}{2.0}} & \textbf{\MS{81.0}{1.1}} & \textbf{\MS{92.8}{1.7}} \\
        \end{tabular}
        \caption{Block3D-Platform Success Rates in percentages in the form \MS{mean}{std-err}, trained over 3 random seeds and evaluated on Lift-Place and Push primitives. PLATO is the only method able to do well on all evaluation primitives and do so consistently, for a wide variety of object dimensions and initial conditions.}
        \label{tab:block2dplat}
    \end{center}
\end{table*}

\smallskip \noindent \textbf{Block3D-Platforms}: Table~\ref{tab:block2dplat} shows the results for Block3D-Platforms. PLATO outperforms the baseline methods in the pushing tasks, but the difference is substantially greater for the lifting tasks. These lifting tasks have longer, variable horizons and involve several bottleneck states (e.g., securely grasping, clearing the platform, dropping on the platform). PLATO is the only method capable of performing well on both the longer horizon tasks (lifting) and shorter horizon tasks (pushing).

\renewcommand{\arraystretch}{1.5}
\setlength{\tabcolsep}{2pt}
\begin{table*}[h]
\fontsize{8.5}{7}\selectfont
    \begin{center}
        \begin{tabular}{c|cccccccccc}
            \multicolumn{1}{c}{} & \textbf{Push-F} & \textbf{Push-L} & \textbf{Push-B} & \textbf{Push-R} & \textbf{Rot(+)} & \textbf{Rot(-)} & \textbf{Lift-F} & \textbf{Lift-L} & \textbf{Lift-B} & \textbf{Lift-R} \\
            \Xhline{2\arrayrulewidth}
            GCBC & \MS{83.7}{1.3} & \MS{91.3}{1.5} & \MS{86.3}{5.4} & \MS{93.0}{2.5} & \MS{78.7}{5.8} & \MS{81.0}{0.6} & \MS{11.1}{6.4} & \MS{27.7}{2.7} & \MS{4.4}{2.5}  & \MS{20.6}{1.3}\\
            
            LMP & \MS{85.0}{2.6} & \MS{87.6}{3.8} & \MS{89.9}{0.6} & \MS{90.6}{3.1} & \MS{51.3}{5.2} & \MS{63.3}{3.2} & \MS{59.3}{5.8} & \MS{74.3}{4.7} & \MS{51.7}{6.6} & \MS{64.8}{5.3} \\
            
            PLATO & \textbf{\MS{94.7}{0.9}} & \textbf{\MS{92.0}{2.1}} & \textbf{\MS{97.2}{1.0}} & \textbf{\MS{98.7}{0.7}} & \textbf{\MS{81.3}{2.5}} & \textbf{\MS{86.3}{6.1}} & \textbf{\MS{81.7}{0.3}} & \textbf{\MS{84.3}{3.0}} & \textbf{\MS{76.7}{10}} & \textbf{\MS{85.2}{3.6}} \\
            \hline 
            LMP(S) & 100 & 100 & 92.7 & 100 & 84.8 & 85.5 & 67.6 & 70.9 & 70.1 & 87.7 \\
            PLATO(S) & 100 & 100 & 97.4 & 100 & 100 & 100 & 92.3 & 90.2 & 82.4 & 85.6 \\
            LMP(S+) & 66.8 & 76.0 & 75.8 & 64.6 & 62.5 & 87.1 & 47.3 & 44.8 & 35.3 & 34.9 \\
            PLATO(S+) & 88.3 & 87.1 & 77.4 & 90.9 & 87.4 & 86.7 & 64.4 & 71.0 &  52.4 & 72.3 \\
            
        \end{tabular}
        \caption{Mug3D-Platform Success Rates in percentages in the form \MS{mean}{std-err}, trained over 3 random seeds and evaluated on Lift-Place, Rotate, Push primitives. In the first block of the table, we see that PLATO is the only method able to do well on all evaluation primitives and do so consistently, for a wide variety of object dimensions and initial conditions. In the second block of this table (generalization experiments, one seed), (S) denotes that the method was trained on a subset of initial mug orientations (simpler tasks). While in principle play will reach other mug orientations, this still skews the distribution of mug orientations towards the initial predefined set. (S+) denotes methods trained on these subset of initial orientations (same models as S) but evaluated on the full range of mug orientations as used in the first block. We see that while LMP performs close to PLATO within distribution (subset of initial mug orientations, significantly less task diversity), PLATO is much more robust than LMP when presented with the full initial object state distribution at test time, showing the generalization capacity of PLATO.}
        \label{tab:mug3dplat}
    \end{center}

\end{table*}

\smallskip \noindent \textbf{Mug3D-Platforms}: Table~\ref{tab:mug3dplat} contains both the results for Mug3D-Platforms presented in the main text, as well as an additional ablation experiment testing the generalization capacity of PLATO in this environment. The main experiments (first block of Table~\ref{tab:mug3dplat}) show that similar to in Block3D-Platforms, PLATO outperforms the methods on all tasks, where the gap is less stark for pushing tasks but especially large on the lifting tasks. For the rotation tasks, interestingly Play-GCBC outperforms Play-LMP. We speculate that in cases like this, the Play-LMP policy might be too dependent on the plan $z$, and thus suffers at test time when the prior outputs an \textit{approximate} $z$ distribution given only partial information. In contrast, Play-GCBC is substantially worse than Play-LMP for the lifting task. This suggests that task variability is much larger for lifting than rotating and pushing, and hence the latent plan in Play-LMP helps the policy disambiguate between this variability. Overall, PLATO performs better and more consistently on all the tasks than either Play-GCBC or Play-LMP.

The second set of experiments (second block in Table~\ref{tab:mug3dplat}) illustrate the robustness of PLATO to state/action/goal distribution shift. We train Play-LMP and PLATO on a subset of initial mug orientations in the mug environment. Specifically, the initial randomized z-axis orientation (yaw) of the mug at the start of each episode will be just a 90 degree cut of the full 360 degree range. While in principle sequential play will be able to eventually see tasks demonstrated for orientations outside this range, this initialization greatly skews the distribution of mug orientations for all demonstrated tasks towards the starting set of orientations. The first two rows of the second block in Table~\ref{tab:mug3dplat} (S) show the performance of Play-LMP and PLATO when evaluated on the tasks used for training (90 degree cut for initial mug orientations). We see here that due to the lower variability in tasks, Play-LMP performs quite well within distribution, notably on the lifting tasks (in contrast to the results from the first block in the table). PLATO performs better than Play-LMP across all tasks, consistent with the results in the first block, although the gap is reduced due to limited task variability. The second two rows of the second block in Table~\ref{tab:mug3dplat} (S+) show the performance when the same models from the first two rows (S) are evaluated on the full swath of initial mug orientations. As explained previously, both Play-LMP and PLATO have seen a limited set of tasks with these orientations due to the sequential nature of play, but only PLATO is able to retain good performance across all tasks. This demonstrates that by biasing the latent space towards learning object affordances, PLATO is better able to capture the full distribution of demonstrated behaviors, even those infrequently demonstrated, and thus is more adept under distribution shift in the test time tasks.

\renewcommand{\arraystretch}{1.5}
\setlength{\tabcolsep}{1.3pt}
\begin{table*}
\fontsize{9}{7}\selectfont
    \begin{center}
        \begin{tabular}{c|cccccccccc}
            \multicolumn{1}{c}{} & \textbf{Push-L} & \textbf{Push-R} & \textbf{Cab-O} & \textbf{Cab-C} & \textbf{Dr-O} & \textbf{Dr-C} & \textbf{To-Cab} & \textbf{To-Dr} & \REV{\textbf{Fr-Cab}} & \REV{\textbf{Fr-Dr}} \\
            \Xhline{2\arrayrulewidth}
            GCBC & \MS{29.1}{3.5} & \MS{52.9}{4.0} & \MS{47.7}{21.4} & \MS{1.2}{1.2} & \MS{86.7}{6.4} & \MS{22.5}{11} & \MS{52.2}{12} & \MS{60.6}{10} & \REV{\MS{6.3}{4.1}} & \REV{\MS{3.7}{0.3}} \\
            
            LMP & \MS{25.5}{2.5} & \MS{60.3}{10.5} & \MS{61.2}{8.3} & \MS{0}{0} & \MS{65.3}{13} & \MS{27.2}{13} & \MS{34.7}{6.2} & \MS{45.5}{7.1}  & \REV{\MS{1.0}{1.0}} & \REV{\MS{19.7}{6.5}} \\
            
            PLATO & \textbf{\MS{76.3}{15}} & \textbf{\MS{66.7}{15}} & \textbf{\MS{100}{0}} & \textbf{\MS{78.7}{6/9}} & \textbf{\MS{100}{0}} & \textbf{\MS{100}{0}} & \textbf{\MS{77.3}{3.7}} & \textbf{\MS{60.4}{2.0}} & \REV{\textbf{\MS{11.7}{1.8}}} & \REV{\textbf{\MS{58.3}{3.5}}} \\
        \end{tabular}
        \caption{Playroom3D Success Rates in percentages in the form \MS{mean}{std-err}, trained over 3 random seeds and evaluated on Push, Cabinet, Drawer, and To\REV{/From}-Cabinet/Drawer primitives. PLATO is the only method able to do well on all evaluation primitives, for a wide variety of object dimensions and initial conditions.}
        \label{tab:playroom3d}
    \end{center}
\end{table*}

\smallskip \noindent \textbf{Playroom3D}: In Table~\ref{tab:playroom3d}, we show the results on Playroom3D. Surprisingly, pushing tasks are much harder in this environment, which we speculate is due to the presence of multiple objects and thus a higher variability in \textit{how} objects can be interacted with. PLATO is able to retain good performance on the pushing tasks, in contrast to the baselines. For opening and closing the cabinet and drawer, we see that PLATO is able to do quite well, but Play-LMP and Play-GCBC perform quite poorly. There is quite a diversity in horizon lengths across these different tasks and different instantiations of each task, which we believe contributes to this large gap. For the Cabinet closing task, there are several critical bottleneck states, for example being able to servo the arm above and to the other side of the cabinet door to reach the cabinet handle, as well as precisely grasping the handle. PLATO is the only method that learns to consistently perform this task. Likewise for the drawer tasks, PLATO gets 100\% success for all random seeds, while performance is substantially worse on the baselines. \REV{We also evaluated two additional challenging tasks in the Playroom3D environment that were demonstrated during play less frequently: From-Cabinet and From-Drawer (pull object out of open cabinet, lift it out of open drawer), also with substantial object and primitive variation. The performance on these tasks are lower compared to the other tasks as they require many pre-conditions and thus are not equally represented during our collected play data. From-Cabinet also requires a novel end effector orientation and grasping procedure in order to avoid collision with the cabinet and table walls. We speculate that due to the novel motion, limited examples, and the presence of other tasks in the data, all methods do notably worse on the From-Cabinet task, however From-Drawer performance for Play-LMP and PLATO is closer to To-Drawer performance since these tasks involve similar object lifting behaviors. However, even with fewer examples in a crowded dataset of other tasks, there is still a large gap between PLATO and the next best method.} Overall, this environment demonstrates that PLATO gracefully scales to more complex environments with more diversity in object affordances and robot behaviors.

\smallskip \noindent \textbf{BlockReal}: Results for our real world experiments are shown in Table~\ref{tab:real}. Both Play-LMP and PLATO are trained entirely in simulation. We add minor data augmentation in simulation in the form of gaussian noise for the object state estimates, to match the observed noise using our real world multi-camera object state estimation infrastructure. Note that both models get 90\%+ success in simulation on each task, since these tasks have relatively low behavior and object diversity. We see that when deployed on the real world setup with no additional data, PLATO experiences only a minor performance degradation, suggesting that learning a latent \textit{affordance} space is more robust than learning a latent \textit{plan} space. These results are consistent with what we see in the Mug3D-Platforms generalization experiments in Table~\ref{tab:mug3dplat}, however here we are testing generalization to entirely unseen real world physics, in contrast to task distribution shift in those experiments.

\subsection{\REV{Additional Analysis}}

\REV{\textbf{Variation in Performance for Similar Tasks}: For several environments, semantically similar tasks like push-left and push-forward seem to have notably different success rates across many methods. Interestingly, those differences are mainly across different ``axes" of the task, for example push-left and push-right usually have similar performance, and push-forward and push-backward also have similar performance, but these two sets have a gap in performance. We speculate that this is because of biases present in the data or the model that favor one axis over another, for example the relative presence of each task, or environmental difficulties in performing that task.}

\REV{\textbf{Quality of Interaction Segmentation}: In all our experiments, we are not assuming access to perfect interaction segmentation. In fact, all of our environments will sometimes have imperfect interaction signals due to the demonstrator having notable noise (even in scripted policies, which have added noise) – for example, brushing against the table, object, or cabinet door on the way to perform a different task. Intermittent contact is also quite common in all of our play data. However, the smoothing on top of the interaction signal tends to clean many of these signals. See the Contact Ablation Experiments in Appendix~\ref{app:detailed_results} for more analysis.}

\REV{Furthermore, we pose this question: what defines ``perfect” segmentation? In the framing of our method, any interaction with the environment, even accidental ones, are still valid for the affordance space to learn. If we accidentally brush the top of the cabinet door on our way to push an object, and the door opens slightly, this can be seen as a successful cabinet slight-open task. Since the start and goal object state are unique for this task, in theory it should not at all affect the affordance learning for a different start and goal object state. However, accidental interactions will start to affect learning if these interactions are common and bias the policy towards unsafe regions of the state (for example, if repeatedly brushing the top of the cabinet door on the way to push the block biases the policy away from pushing the block properly).}

\renewcommand{\arraystretch}{1.25}
\setlength{\tabcolsep}{4pt}
\begin{table*}
    \begin{center}
        \begin{tabular}{c|cccc}
            \multicolumn{1}{c}{} & \textbf{Push-Left} & \textbf{Push-Back} & \textbf{Push-Right} & \textbf{Push-Forward} \\
            \Xhline{2\arrayrulewidth}
            LMP & 8/10 & 6/10 & 8/10 & 7/10 \\
            PLATO & 8/10 & \textbf{8/10} & \textbf{9/10} & \textbf{10/10} \\
        \end{tabular}
        \caption{Results for BlockReal on pushing tasks. These models are trained entirely in simulation with minor data augmentation. We evaluate these models on a real robot setup, and see that performance degrades less for PLATO than for Play-LMP when presented with the real world object and robot dynamics. Methods that use play data are robust to environment changes, consistent with results from prior work~\cite{lynch2020learning}.}
        \label{tab:real}
    \end{center}
\end{table*}

\section{Additional Limitations}
\label{app:limits}

\REV{We will now present a longer discussion of limitations presented in Sec.~\ref{sec:conclusion}, as well as some additional limitations, to help guide future work.}

\REV{\textbf{Multi-Object Scenarios}: As noted in Sec.~\ref{sec:conclusion}, multi-object interaction scenarios like tool-use represent a key challenge for future work. While we show PLATO operating in multi-object environments in this work, we do not extend PLATO to multi-object \textit{interactions} involving dynamic objects, for example hitting a puck with a hockey stick. In these settings, it might be difficult to obtain signals for interaction between the dynamic objects. We will give a couple of ideas of how future work might tackle this problem in the hopes of opening up a broader discussion. Before these ideas, one general point of clarification: PLATO introduces detecting ``interaction" as a more general concept, which applies even when no contact occurs between the robot and the desired object (e.g., tool use). We used contact as our interaction signal since it was the most readily available for single-object interactions. However for multi-object interactions, the notion of interaction still exists and our method still applies if we can detect these interactions somehow. Since PLATO only requires detecting interaction during training, one option is to have a human label interaction segments in their training data manually. If this is too time intensive, we can potentially leverage \textit{learned} binary signals for interaction using limited supervised interaction labels. Consider the tool-use task of hitting a hockey puck with a hockey stick. Even though we cannot directly observe a contact signal between the stick and the puck, future work could learn to predict the interaction signal from visual (e.g. observing the stick hit the puck) and maybe even haptic information (force feedback of the hockey stick on the end-effector when hitting), using supervised labels. In addition, recent approaches like ComPILE~\cite{kipf2019compile} have learned to segment skills without any labels, and could be used to isolate an interaction signal for dynamic multi-object interactions in a self-supervised fashion. While designing such a system for detecting interaction might require some effort, we believe that PLATO’s results suggest that such effort can result in serious performance gains for policy learning from play.}

\REV{With this general notion of interaction in mind, we would like to present two potential ideas for future work to expand PLATO to multi-object scenarios:}
\REV{\begin{enumerate}
    \item One idea would be to learn an \textit{embodiment-specific} affordance space. Then, each tool can be viewed as a different embodiment of the robot, and with knowledge of what tool is currently being used, we can learn affordances specific to that tool. At test time if we know what tool we picked up, the policy can leverage the affordance space of this particular tool in a manner similar to PLATO. 
    \item As another related idea, we could attempt to learn multiple \textit{degrees} of interaction: e.g. similar to how PLATO learns the relationship between an object affordance (hockey stick moves) and robot skill (grasp and swing a stick), we might also learn the relationship between a second order object affordance (hockey puck moves) and a first order object affordance (hockey stick strikes). Then at test time the policy could reason backwards in time, first inferring the correct second order affordance (hockey puck moves), then the first order affordance (hockey stick swings), then the robot action to take (grasp and swing the stick).
\end{enumerate}} 

\REV{We see our work as a first step towards exploring some of these interesting paradigms for interaction and multi-level skill reasoning.}

\REV{\textbf{Regularization Weight}}: As one might expect, both Play-LMP and our algorithm PLATO share many of the challenges of variational auto-encoders~\cite{pmlr-v33-ranganath14}. Recent replications of this work have shown that the regularization weight $\beta$ has a sizeable effect on the final policy reconstruction error~\cite{douglas_2021}. High values of $\beta$ can yield posterior collapse of the latent space, where the plan posterior outputs distributions for differing trajectories that do not reflect these differences; low values of $\beta$ can yield low reconstruction losses, but conversely cause the posterior plans to encode information that is hard to predict from just the start and end states. As a result, the learned prior may not match plans encoded by the posterior, thus hurting the policy. Future work might look into more expressive learned priors, such as mixture models, in order to better match the posterior and thereby reduce the sensitivity to $\beta$. Another direction could be finding alternate ways to specify tasks at test time, for example giving some notion of \textit{how} a goal should be reached.

\REV{\textbf{Human Sub-Optimality}}: Additionally, when collecting play data, certain challenging primitives attempted by humans might fail often. We noticed that humans often fail at Side-Rotate in Block2D, for example, and the resulting demonstration might look like a sub-optimal Push from the perspective of the posterior and the prior. This introduces even more plan variability into the latent space for the Push task, and thus hurts test time performance. 
Another interesting direction would be to better understand how sub-optimality affects the learned latent space and potentially develop a notion of trajectory ``quality" to bias this latent space.

\REV{\textbf{Use of Object State Estimation}: As mentioned in Sec.~\ref{sec:conclusion}, PLATO makes use of object state estimates when learning object-centric affordances. A natural question is how this work can be adapted to work with pure image inputs. Critically, the only obstacle to extending our method to image inputs is our learned prior network, which utilizes just ground-truth object state information rather than the full proprioceptive state and object state. Note that our PLATO-R ablation can be extended to learn from images without any additional modifications, however this ablated method lacks the benefits of object-focused affordance learning (see Section~\ref{sec:exp}). While we did not explore learning from images, we believe our method can readily scale to images with a few modifications. One method would be to mask out only the object(s) of interest from the start and goal images before passing them into the prior to encourage a robot agnostic latent space. Another method would be to learn an object representation directly from images that is independent of the robot state (e.g. with contrastive learning on negative examples of different robot poses, but identical environment states).}

\REV{Regardless, we showed that learning object centric representations actually improves the robustness of policies at test time, and thus justifies the extra effort of designing these representations. Furthermore, there is a sizable research field devoted to improving object state estimation using learning and filtering techniques (even involving state estimation under clutter), so we believe that object state estimation methods will get even more practical in the coming years.}

\end{document}